\documentclass{article}

\usepackage[preprint]{neurips_2026}

\usepackage[utf8]{inputenc} 
\usepackage[T1]{fontenc}    
\usepackage{hyperref}       
\usepackage{url}            
\usepackage{booktabs}       
\usepackage{amsfonts}       
\usepackage{nicefrac}       
\usepackage{microtype}      
\usepackage{xcolor}         
\usepackage{amsmath}
\usepackage{multirow}
\usepackage{enumitem}
\usepackage{amssymb}
\usepackage{graphicx}
\usepackage{tikz}
\usetikzlibrary{arrows.meta,positioning}
\usepackage{upquote}   
\usepackage{listings}
 \usepackage{array}
 \newcolumntype{L}[1]{>{\raggedright\arraybackslash\hyphenpenalty=10000\exhyphenpenalty=10000}p{#1}}
\hypersetup{hidelinks}

\title{Evaluating Agents Across Runtime Contracts:\\ When Mismatch Costs Efficiency or Quality}

\workshoptitle{NeurIPS 2026 Workshop on Evaluation of Interactive Agents}
\author{
  Victor May$^{1}$, Van Khue Nguyen$^{2}$, Aaditya Salgarkar$^{1}$, \\
  \bfseries Yishan Wang$^{3}$, Diganta Misra$^{4,5,6}$, Huu Nguyen$^{1}$ \\
  \rule{0pt}{2em} \normalfont $^1$Ontocord \quad $^2$École Polytechnique \quad $^3$Carnegie Mellon University \\
  \normalfont $^4$MPI-IS Tübingen \quad $^5$Tübingen AI Center \quad $^6$ELLIS Institute Tübingen
  \rule{0pt}{2em} \normalfont 
}

\begin{document}

\maketitle

\begin{abstract}
Language-model agents can solve tasks by writing Python code, executing tool calls within it, and
using the feedback to choose their next action. In this CodeAct setting, the execution runtime may
preserve Python variables between actions or reset them after each one, without resetting task
progress. Training on execution traces therefore demonstrates not only how to solve a task, but also
how intermediate results can be stored and reused. We study \emph{runtime transfer}: whether agents
trained under one execution contract remain effective under the other, a dependence that evaluations
under a fixed runtime can conceal. Across three procedural task families requiring agents to build and
reuse records of tool feedback, we generate paired training traces for the same instances under
persistent and stateless runtimes. We fine-tune separate Qwen3-8B agents for each task and runtime,
evaluate all four training--deployment combinations, and vary the per-turn tool-call cap: the number
of tool calls one Python action may execute. In our primary task, agents inspect items with hidden
attributes, build a record of those inspections, and select a high-value subset under a weight limit.
At $80$ calls per turn, a persistent-trained agent deployed statelessly achieves
$0.61$ of optimal value, compared with $0.77$ under matched
persistent deployment, but uses $4.4\times$ as many tokens while repeatedly
reconstructing its working record. Tightening the cap to $25$ reduces mismatched
quality to $0.07$, while matched quality remains
$0.66$. Trace replay shows how the cap interrupts an inspection loop, the reset
erases its partial record, and rebuilding it consumes the calls needed for further progress. This
amplification of the runtime gap replicates across training seeds, an independent second rollout, and
two additional base models. Across the tasks and controls studied, mismatch-specific quality collapse
appears only when the cap interrupts an intended action and matched execution can resume from
surviving state while mismatched execution restarts. Evaluations should therefore test runtime
transfer and report and vary runtime contracts, per-turn caps, and episode horizons together,
measuring operational cost alongside task quality. Otherwise, the same mismatch may appear mainly as
redundant computation under one configuration and as near-total quality loss under another.
\end{abstract}

\section{Introduction}\label{sec:intro}
Tool-using language-model agents solve tasks by alternating between reasoning, actions and feedback
from an environment \citep{yao2023reactsynergizingreasoningacting}. In the CodeAct paradigm the action
is executable code: The model emits a Python block, an interpreter runs the block and the tool calls
inside it, and the output or error comes back as the next observation
\citep{wang2024executablecodeactionselicit}. Benchmarks such as InterCode, MINT, $\tau$-bench and
ToolSandbox evaluate agents through related multi-turn loops
\citep{yang2023intercodestandardizingbenchmarkinginteractive,wang2024mintevaluatingllmsmultiturn,yao2024taubenchbenchmarktoolagentuserinteraction,lu2025toolsandboxstatefulconversationalinteractive}.

Three terms describe that setup. The \emph{interpreter} runs the agent's Python code and holds the
variables the code creates. The \emph{runtime contract} fixes what happens to those variables between
turns: Under a \emph{persistent} runtime, a variable created on one turn is still bound on the next,
while under a \emph{stateless} runtime, every turn starts from a fresh namespace. If one action stores
inspected item records in a variable \texttt{items}, the next action can reuse \texttt{items} only
under the persistent contract. Task and tool state survive under both, so a stateless runtime resets
the agent's Python bindings but not the task. The \emph{harness} is the system around all of this: It
parses the action, calls the interpreter and the tools, formats the observation, and enforces limits
such as the number of turns or tool calls.

Those benchmarks hold the harness fixed and compare agents inside it. That is a natural design, but it
makes task skill hard to separate from compatibility with one particular interface.
Agent-computer interface design already moves measured performance
\citep{yang2024sweagentagentcomputerinterfacesenable}, and recent work measures sensitivity to harness
configuration across models and workflows
\citep{yao2026harnessbenchmeasuringharnesseffects,cho2026itscapabilityharnesssensitivity}. A score
belongs to the agent and to the harness it ran in.

For agents fine-tuned on interaction traces, that compatibility is partly learned. Trace
fine-tuning is now a common way to train tool-using agents
\citep{chen2023fireactlanguageagentfinetuning,zeng2023agenttuningenablinggeneralizedagent,song2025agentdataprotocolunifying}.
A persistent trace demonstrates that maps, records and helper functions can be left in variables
and picked up later. A stateless trace demonstrates that the same information has to be printed,
rebuilt from the transcript or fetched again through tool calls.

We call the resulting evaluation question \emph{runtime transfer}: Does an agent trained under
one runtime contract still solve the task when deployed under the other? The question is not which
runtime is easier. It is whether the agent learned a strategy that depends on the contract demonstrated in its
training data and what happens when deployment breaks that contract.

To make runtime transfer observable, we use tasks that require the agent to build a structured working
record from tool feedback and to use that record in later actions (Sec.~\ref{sec:design}). We study
three such families, defined in App.~\ref{app:tasks}. In Opaque Knapsack, the record is a table of
inspected items, in Navigation, a map of locks, traps and keys, and in Rule Diagnosis, a set
of probes and a hypothesis revised from counterexamples. Each family stays solvable under either
runtime, so persistence never decides whether the task can be solved. Instead, it changes how the agent has
to organize its computation.

\begin{figure}[tb]
\centering
\includegraphics[width=\linewidth]{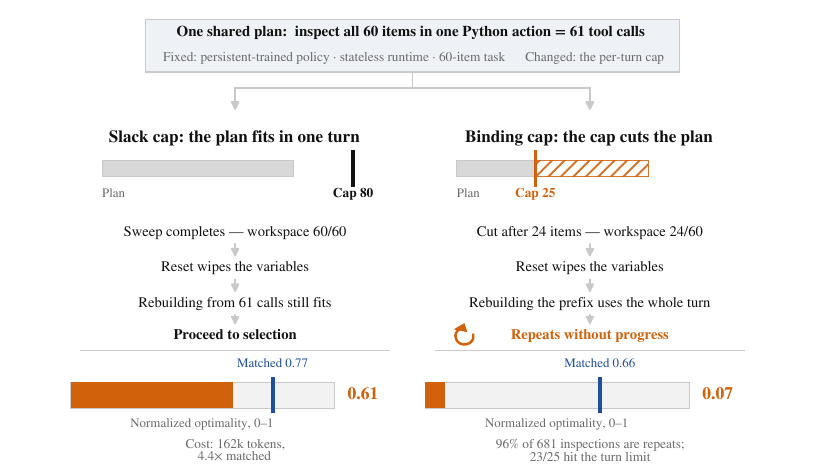}
\caption{\textbf{The same runtime mismatch costs computation under a slack cap and quality under a
binding cap.} We fix a persistent-trained agent on the stateless runtime and vary the per-turn cap. At
$c{=}80$ the complete $61$-call sweep drawn here fits in one
turn, so the agent rebuilds after the reset and retains much of matched quality at
$4.4\times$ the token cost. At
$c{=}25$ the cap truncates the sweep, the reset erases the partial working record,
and rebuilding the prefix consumes the next turn.
Bars are mean normalized optimality over $n{=}25$ paired tasks
(Table~\ref{tab:main}). Blue markers are matched persistent deployment. A representative excerpt is in
App.~\ref{app:cap}.}
\label{fig:hero}
\end{figure}

For every training instance, we generate one accepted training trace under each runtime
contract. We fine-tune a separate agent on the persistent and on the stateless traces, then
evaluate both agents under both runtimes. That gives four conditions: two matched, where training and
deployment share a contract, and two transfer conditions, where they differ. Holding the trained agent
and the held-out instances fixed, we then vary the per-turn tool-call cap $c$, the largest number of
tool calls one Python action may execute. The cap stresses the strategy each agent learned for storing and recovering
its working record.

The cap decides how the mismatch appears. Under a slack cap ($80$ calls per
turn, which the agent's action blocks rarely reach), a persistent-trained agent on the stateless
runtime keeps much of matched quality
($0.61$ of optimal against
$0.77$), but spends
$4.4\times$ as many tokens rebuilding
records the persistent runtime would have kept. Under a binding cap ($25$, which
routinely truncates them) the same agent falls to $0.07$, while matched deployment stays at
$0.66$. The cap interrupts the inspection loop, the reset that
follows removes the partial record, and the agent restarts instead of continuing
(Fig.~\ref{fig:hero}). An evaluation at a single slack cap sees substantial reconstruction overhead but
misses the much larger quality loss a tighter cap exposes. Our contributions:

\noindent\begin{minipage}{\linewidth}
\begin{itemize}[leftmargin=1.4em,itemsep=1pt,topsep=1pt]
\item \textbf{A controlled evaluation of runtime transfer.}
We train agents on persistent and on stateless traces, then evaluate both under
both runtimes, on three tasks that require a working record built from tool feedback.
\item \textbf{A replicated cap-dependent failure mode.}
Tightening the per-turn cap changes the cost of mismatch from repeated
reconstruction to severe quality loss, across a dense cap sweep and replicated
across training seeds, a second rollout and two additional base models.
\item \textbf{A behavioral diagnostic of cap-amplified mismatch.}
Across the settings studied, tightening the cap collapses quality under
mismatch only where it interrupts an intended action and matched deployment can
resume from surviving state. Preserving variables is not enough on its own: the
agent must also know how to resume.
\end{itemize}
\end{minipage}

\section{Related work}

\paragraph{Interactive agent evaluation and harness sensitivity.}
The multi-turn benchmarks named in Sec.~\ref{sec:intro} score interaction with execution or
environment feedback rather than one-shot answers, and each compares policies within one fixed
agent-computer interface
\citep{yang2023intercodestandardizingbenchmarkinginteractive,wang2024mintevaluatingllmsmultiturn,yao2024taubenchbenchmarktoolagentuserinteraction,lu2025toolsandboxstatefulconversationalinteractive}.
Both interface design and harness configuration move the resulting scores
\citep{yang2024sweagentagentcomputerinterfacesenable,yao2026harnessbenchmeasuringharnesseffects,cho2026itscapabilityharnesssensitivity}.
We study a complementary form of sensitivity: what happens when the runtime contract at deployment
differs from the one the agent's training traces may have taught it to expect. Our four-condition design varies the training and
deployment runtime separately, while the per-turn tool-call cap provides a controlled stress test of
that transfer.

\paragraph{Fine-tuning on interaction traces and learned runtime assumptions.}
Executable actions are one of several ways to interleave language-model reasoning with tool use
\citep{gao2023palprogramaidedlanguagemodels,chen2023programthoughtspromptingdisentangling,gou2024toratoolintegratedreasoningagent,yao2023reactsynergizingreasoningacting,wang2024executablecodeactionselicit},
and agent abilities are increasingly trained from the resulting interaction traces
\citep{chen2023fireactlanguageagentfinetuning,zeng2023agenttuningenablinggeneralizedagent,song2025agentdataprotocolunifying}.
Those traces convey more than task-relevant actions: they also demonstrate properties of the
interface under which those actions succeeded, including where intermediate state can be stored and
whether it remains available on later turns. We study these learned runtime assumptions by separating
the runtime demonstrated during fine-tuning from the runtime enforced at evaluation, and by varying an
evaluator-controlled per-turn cap to determine when runtime mismatch costs efficiency or task quality.

\paragraph{State management and budget-constrained tool use.}
Work on long-horizon agent memory studies how intermediate state can be retained, compressed or
reconstructed through managed context
\citep{packer2024memgptllmsoperatingsystems,wu2026resumunlockinglonghorizonsearch,zhou2025mem1learningsynergizememory}. A separate line studies how
agents allocate limited tool calls and adapt to call budgets
\citep{liu2025budgetawaretooluseenableseffective,liu2026costbenchevaluatingmultiturncostoptimal,ding2026calibratethenactcostawareexplorationllm}.
Our setting connects these questions but differs from both. The state-management strategy is partly
induced by the training runtime, while the cap is enforced by the evaluator at each action boundary
rather than supplied as an episode-level budget for the agent to optimize. The cap therefore serves as a
diagnostic: it reveals when recovering state learned under one runtime consumes the capacity needed to
make new progress under another.

\section{Runtime-transfer evaluation design}\label{sec:design}
Our evaluation separates the runtime contract demonstrated during training from the one used at
deployment: generate paired teacher traces under both runtimes, train one adapter on each,
cross-evaluate both under both runtimes, then vary the per-turn tool-call cap within every condition. It asks
whether a policy learned under one runtime transfers to the other, and whether a setting the evaluator
controls changes the cost of that transfer.

\subsection{Runtime contracts and the four conditions} A turn is one attempted model call and executes
at most one Python action block. Failed calls and rejected
completions spend a turn of the horizon without executing a block
(App.~Table~\ref{tab:horizon}). We use \emph{action block} for the
unit a cap truncates and \emph{turn} for the unit the episode horizon counts. Our two interpreter
semantics differ only in whether agent-created bindings survive the turn boundary. Task and
environment state, including the tool objects, persist under both. Each runtime condition also
announces itself to the agent, through a mode instruction, a runtime-state banner and
runtime-specific few-shot examples (App.~\ref{app:runtime}). Cross-runtime comparisons therefore
identify the effect of switching this package, not interpreter persistence in isolation.
Cross-adapter comparisons additionally bundle differences in training-trace organization and length.

\begin{table}[h]
\centering\small
\caption{\textbf{The four conditions.} We write $X\!\to\!Y$ for an $X$-trained agent deployed on a
$Y$ runtime, $X,Y\in\{P,S\}$.}
\label{tab:design}
\begin{tabular}{@{}lll@{}}
\toprule
Training runtime & Persistent deployment & Stateless deployment\\
\midrule
Persistent & $P\!\to\!P$: matched & $P\!\to\!S$: main mismatch\\
Stateless & $S\!\to\!P$: reverse transfer & $S\!\to\!S$: matched\\
\bottomrule
\end{tabular}
\end{table}

Each task family has two trained agents: one trained on persistent traces and one trained on stateless
traces. We evaluate each agent under both runtimes, producing the four conditions in
Table~\ref{tab:design}. The two matched conditions are the references, and the two transfer conditions
are the experiment. We focus on $P\!\to\!S$, where the agent learned that variables would persist but
the deployment runtime removes them after each turn. The reverse condition, $S\!\to\!P$, controls for
the general effect of changing runtimes.

\subsection{Tasks and paired trace training}
Beyond the requirements in Sec.~\ref{sec:intro}, a usable task must score rebuilt work apart from new
progress, so that a rebuild registers as cost and not as credit. Table~\ref{tab:tasks} lists the three
procedural families that meet all of these requirements. Opaque Knapsack is the primary evaluation, and its
inspection phase is deliberately \emph{batchable}: under a large enough cap the whole item table fits
in one block, so the cap alone decides whether that computation crosses a turn boundary.
Navigation and Rule Diagnosis are two smaller control tasks. Their records are revised as feedback
arrives rather than accumulated in a single sweep, so the cap acts on them differently.
App.~\ref{app:task_design} gives the generators and anti-degeneracy constraints,
App.~\ref{app:task} the Knapsack tool contract and instance distribution, and App.~\ref{app:crossfam}
the controls.

\begin{table}[tb]
\centering\small
\caption{\textbf{Three tasks built so that the runtime matters.} Each needs a record assembled from
tool feedback that later actions depend on, and each is solvable under either runtime.}
\label{tab:tasks}
\begin{tabular}{@{}L{0.195\linewidth}L{0.34\linewidth}L{0.40\linewidth}@{}}
\toprule
Task & Working record & What a reset disrupts\\
\midrule
Opaque Knapsack & Table of inspected items & A reset erases the table left by a cut\\
Navigation & Map of locks, traps and keys & A reset erases a partly checked map\\
Rule Diagnosis & Probes and a revised hypothesis & A reset breaks the gathered evidence\\
\bottomrule
\end{tabular}
\end{table}

Each family has its own adapter pair, and none is trained on a mixture of families or run on one it
did not see. Traces are paired by instance: an instance is retained only if both of its teacher
traces independently pass the same validity, quality and context-length filters. Each family retains
$1{,}000$ paired instances per training runtime, the two sides
carrying identical task text and tool schemas. We fine-tune Qwen3-8B
\citep{yang2025qwen3technicalreport} on each set with QLoRA
\citep{dettmers2023qloraefficientfinetuningquantized}. App.~\ref{app:train} gives the filters and the
training configuration. Opaque Knapsack teacher traces are generated at $c{=}25$, so the binding-cap
evaluation introduces no change in the per-turn cap between training and deployment, and the slack cap
is the setting that departs from it. Pairing controls
which instances are used but not the traces they produce: stateless traces rebuild their state
and run $\approx2.3\times$ longer (Sec.~\ref{sec:limitations}). The cap
sweep within a fixed condition holds all of that constant.

\subsection{Varying the cap, and the three quantities we report}
\textbf{Cap intervention.} The per-turn cap $c$ is the only limit the intervention changes. It leaves
the task's probe and inspect budgets and the episode's turn, context and time limits alone
(Table~\ref{tab:limits}). One turn may still contain many calls, since the emitted block can call
tools inside loops.

\begin{table}[h]
\centering\small
\caption{\textbf{The limits an evaluation sets, and what each one controls.} The intervention changes
the first row only. A failed or rejected model call spends a turn of the horizon without executing an
action block, so attempted turns and executed blocks are not the same count
(App.~Table~\ref{tab:horizon}).}
\label{tab:limits}
\begin{tabular}{@{}ll@{}}
\toprule
Limit & What it controls\\
\midrule
Per-turn cap $c$ & Tool calls inside one action block\\
Task budget & How much hidden task information the agent may reveal\\
Turn limit $T_{\max}{=}40$ & Model calls an episode may attempt\\
Token and context limit & Maximum transcript length\\
Wall-clock limit & Maximum execution time\\
\bottomrule
\end{tabular}
\end{table}

When an action reaches the cap, the runtime stops it before the next tool call. Output printed earlier
in the action remains visible, and the rejected call does not consume the task budget. The variables
created before the interruption then follow the runtime contract, so the cap decides whether the agent
can finish a planned action before a reset.

We preselected $c{=}25$ and $c{=}80$ as the binding and slack
settings from the task's $\approx\!50$-observation reconstruction phase,
before the main evaluation ran, though not formally pre-registered. Those labels name intended regimes. They do not
assume that every policy reaches one cap. Whether a cap is reached in practice is measured from the
realized traces: by the cap-hit rate, the mean per-episode fraction of logged steps that reach the
cap,
and, where a slack-cap rollout exists, by the fraction of blocks a smaller cap would have truncated.
Sec.~\ref{sec:main} reports those rates. Sec.~\ref{sec:mech} shows why nominal task demand does not
predict them.

\textbf{Outcomes and quantities.} The primary evaluation uses the same
$25$ held-out instances of $40$--$65$ items in every
condition and at both caps, and their first $8$ for the extra seeds. Every
episode runs at $T_{\max}{=}40$ turns. We score by \emph{normalized optimality}: the
agent's selected value divided by the exact optimum (App.~\ref{app:task}). App.~\ref{app:train} gives
decoding and serving settings. Let
$Q_{XY}(c)$ be mean quality for an adapter trained under runtime $X$ and evaluated under runtime $Y$,
$X,Y\in\{P,S\}$. We report three related quantities, each answering a different question. The
runtime gap $G_P(c)$ asks how much quality the persistent-trained agent loses when it moves to the
stateless runtime. The interaction $D(c)$ asks whether persistent rather than stateless deployment
benefits the persistent-trained agent more than the stateless-trained agent. Cap moderation $T$ asks whether that interaction
grows when the cap tightens from $80$ to $25$.
\[
\begin{aligned}
G_P(c) &= Q_{PP}(c)-Q_{PS}(c) && \text{runtime gap},\\
D(c) &= \bigl(Q_{PP}(c)-Q_{PS}(c)\bigr)-\bigl(Q_{SP}(c)-Q_{SS}(c)\bigr)
  && \text{train}\times\text{runtime interaction},\\
T &= D(25)-D(80) && \text{cap moderation}.
\end{aligned}
\]
Applying the same tightening to the runtime gap itself gives
$G_P(25)-G_P(80)$, which we call the \emph{cap
amplification}. It is the quantity Fig.~\ref{fig:arms} reports.

Intervals are paired on the task instance: seed-0 intervals resample held-out tasks, the pooled
multi-seed interval for $T$ uses a crossed seed$\times$task bootstrap \citep{efron1993bootstrap}, and we
treat the sign replication of the three per-seed $T$ values as the primary multi-seed evidence
(App.~\ref{app:stats}). Each condition is one temperature-$0.2$ rollout per task with no
decoding seed. Intervals resample the observed task-level outcomes, so with one rollout per task and
condition they do not separately estimate within-task decoding variability. An independent second
rollout assesses sensitivity to it in Sec.~\ref{sec:main}.

\section{A binding cap turns reconstruction overhead into task failure}\label{sec:main}
On Opaque Knapsack, the persistent-trained agent stays near matched quality for as long as its
reconstruction fits inside one turn: at the slack cap ($c{=}80$), it scores
$Q_{PS}{=}0.61$ against
$Q_{PP}{=}0.77$. It pays in redundant work instead, at
$162$k tokens per episode,
$4.4\times$ the matched condition's. Token counts are the usage recorded for accepted completions, and
App.~\ref{app:tmax} audits what that leaves out: neither cell behind this ratio discarded a single
completion. It also spends
$3.8$ turns per episode raising a missing-variable error (\texttt{NameError}) on
variables the reset wiped. At the binding cap ($c{=}25$), it collapses to
$0.07$ while matched deployment holds at
$0.66$ (Table~\ref{tab:main}).

\begin{table}[tb]
\centering\small
\caption{\textbf{The $2\times2$ experiment at binding and slack per-turn caps.} Entries are mean normalized
optimality over the same $n{=}25$ paired instances (seed-0).
$G_P=Q_{PP}-Q_{PS}$ carries a task-paired bootstrap $95\%$ interval. Each row
carries both training regimes' runtime gaps, so the train$\times$runtime interaction $D$ is the
difference between them. The last two columns summarize operational cost in the $P\!\to\!S$ condition.
Per-condition intervals and the termination taxonomy are in App.~Table~\ref{tab:main2x2}.}
\label{tab:main}
\setlength{\tabcolsep}{4.5pt}
\begin{tabular}{lccccccc}
\toprule
& \multicolumn{2}{c}{$P$-trained} & \multicolumn{2}{c}{$S$-trained} & & \multicolumn{2}{c}{$P\!\to\!S$ cost}\\
\cmidrule(lr){2-3}\cmidrule(lr){4-5}\cmidrule(lr){7-8}
Cap & $P\!\to\!P$ & $P\!\to\!S$ & $S\!\to\!P$ & $S\!\to\!S$ & Runtime gap $G_P$ & Tokens/ep. & Cap-hit \\
\midrule
$c{=}25$ (binding) & 0.66 & $\mathbf{0.07}$ & 0.12 & 0.26 & +0.59 [+0.46,+0.71] & 498k & 71\%\\
$c{=}80$ (slack) & 0.77 & 0.61 & 0.23 & 0.51 & +0.15 [+0.00,+0.32] & 162k & 4\%\\
\bottomrule
\end{tabular}
\end{table}

\textbf{The failure is not caused by too few calls overall.} The episode permits up to
$1{,}000$ calls, far more than the roughly
$50$ observations the task requires. The matched agent still scores
$Q_{PP}{=}0.66$ under the same cap, in
$9.8$ logged steps, from
$13.2$ attempted turns (App.~Table~\ref{tab:horizon}). The cap binds mainly at $c{=}25$:
$15$--$71\%$ of logged steps reach it
there against $\le4\%$ at
$c{=}80$.

\begin{figure}[tb]
\centering
\includegraphics[width=0.495\linewidth]{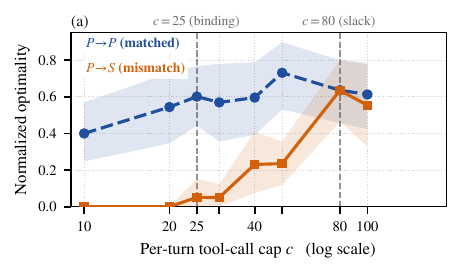}\hfill
\includegraphics[width=0.495\linewidth]{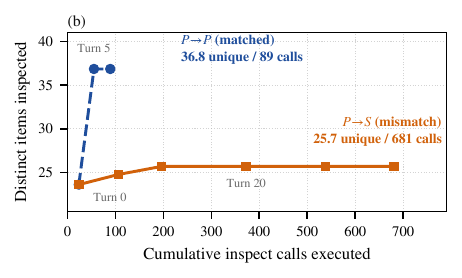}
\caption{\textbf{Cap tightening selectively harms the mismatched condition.} (a) Mean normalized optimality
across $8$ caps on $n{=}12$ paired tasks, unsmoothed, with
paired bootstrap $95\%$ intervals. The matched condition, swept on the same tasks, does
not collapse. (b) Distinct items gained after turn 0 against cumulative \texttt{inspect} calls at
$c{=}25$ on the main evaluation's $n{=}25$ paired tasks, measured from a common turn-0 baseline of
$23.6$ items on average: the mismatched agent
issues $681$ calls to add only
$2.1$ items, where matched
deployment adds $13.2$ in
$89$, over the same instances.}
\label{fig:dose}
\end{figure}

The persistent-trained agent on the stateless runtime spends its calls on reconstruction rather than
on new progress. It repeatedly starts reconstruction from the item listing, with
$30.8$ listing-call occurrences per episode including the
initial listing, and exhausts the turn horizon
in $23$ of $25$ episodes. It issues
$681$ \texttt{inspect} calls but reaches only
$25.7$ distinct items, so $96\%$ of those
inspections are repeats (Fig.~\ref{fig:dose}b).

Scored after the episode with the exact solver, the inspected set already contains a
$0.71$-of-optimal solution, against
$0.83$ for matched deployment (App.~\ref{app:replay}). The agent realizes
only $0.09$ of that oracle-within-set value, against
$0.80$ matched. The much larger loss therefore arises after item
acquisition, in admissibility inference, selection and execution, rather than from item coverage
alone. This persistent-trained policy keeps its record almost entirely by reissuing tool
calls, so once repeated
inspection fills each action block, further turns produce repetition rather than progress.

A larger episode horizon and wall-clock limit do not rescue the condition within the tested context
limits. Quadrupling the horizon leaves $P\!\to\!S$ flat and collapsed while the matched control remains stable
(App.~\ref{app:tmax}). Only
$3/25$ episodes overflow at
$T_{\max}{=}40$, so context overflow cannot explain the collapse of the whole condition.

\textbf{The result appears across the full cap sweep.} Across a dense sweep of
$8$ caps from $c{=}10$ to $c{=}100$,
$P\!\to\!S$ quality improves with $\log c$ ($\beta_{PS}{=}+0.30$
$[+0.23,+0.37]$). The matched condition, swept over
the same range on the same tasks, stays comparatively flat ($+0.10$
$[-0.01,+0.19]$). The paired slope difference is
$\beta_{PS}-\beta_{PP}=+0.20$
$[+0.09,+0.31]$, which excludes $0$
(Fig.~\ref{fig:dose}a, Table~\ref{tab:dense}). The rise is not strictly monotone: the sweep peaks at
$0.64$ at $c{=}80$ and dips to
$0.55$ at $c{=}100$, within overlapping intervals. That is
why we summarize by the continuous slope rather than by any single pairwise comparison. The sweep is a separate
$n{=}12$ evaluation on paired task IDs, and it agrees with the main evaluation at
$c{=}25$ and $c{=}80$ (App.~\ref{app:dense}).

\begin{figure}[tb]
\centering
\includegraphics[width=0.75\linewidth]{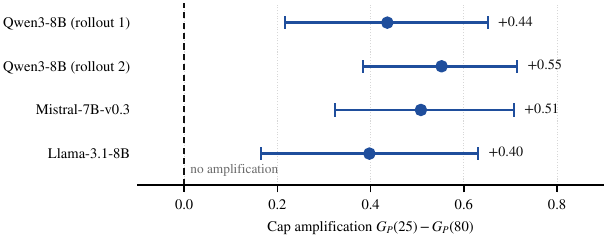}
\caption{\textbf{The runtime gap increases when the cap is tightened from
80 to 25 in all four replications.} Points are
$G_P(25)-G_P(80)$ over the same
$25$ instances, and bars are paired bootstrap
$95\%$ intervals. Replications 1--2 are two rollouts of the same adapters and differ only
in sampling. Replications 3--4 retrain the same recipe on Mistral-7B-v0.3 and Llama-3.1-8B. The full
factorial, the model-selection difference and the scope of each replication are in
App.~\ref{app:arms}.}
\label{fig:arms}
\end{figure}

\textbf{The larger gap at the binding cap replicates across models and rollouts.}
Fig.~\ref{fig:arms} reports the cap amplification: how much the persistent agent's runtime gap
increases when the cap tightens from $80$ to $25$. It is
positive in both Qwen3-8B rollouts and, after retraining, the same recipe on
Mistral-7B-v0.3 \citep{jiang2023mistral7b} and on Llama-3.1-8B
\citep{grattafiori2024llama3herdmodels}. It ranges from
$0.40$ to
$0.55$ over the same $25$ instances, and
all four task-paired $95\%$ intervals exclude zero. The full train$\times$runtime
interaction grows as well, from $+0.43$ at
$c{=}80$ to $+0.73$ at
$c{=}25$, giving $T{=}+0.31$
$[+0.04,+0.57]$ on the main
evaluation ($n{=}25$). We observe the same direction in all three training
seeds (App.~\ref{app:stats}).

These replications cover the Opaque Knapsack factorial only (App.~\ref{app:arms}). The other tasks'
controls and the boundary interventions use the main Qwen3-8B setting.

\section{When cap tightening amplifies runtime mismatch: a two-condition diagnostic}\label{sec:mech}
\begin{figure}[tb]
\centering
\includegraphics[width=\linewidth]{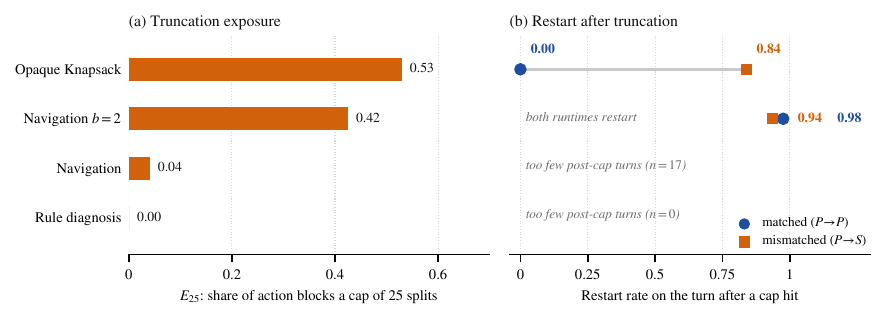}
\caption{\textbf{In the four settings studied, cap tightening amplifies mismatch only where truncation
is common \emph{and} matched execution resumes.} Same four settings, same rows in both panels. (a)
Exposure $E_{25}$ for the mismatched $P\!\to\!S$ condition: the share of its slack-cap
action blocks a cap of $25$ would split. (b) Post-cap restart rate under matched and mismatched
deployment, meaning the agent re-runs its listing call before acquiring anything new. Knapsack meets both
conditions. With batch-limited Navigation both runtimes restart, so both deteriorate and we do not
detect additional mismatch-specific amplification. Plain Navigation
($17$ post-cap turns) is too sparse
for (b), and Rule Diagnosis has none (App.~\ref{app:diag}).}
\label{fig:gates}
\end{figure}

Why does the same cap interrupt actions in one task but not in another? Across the four settings
studied, tightening the cap produces a pronounced mismatch-specific quality collapse only where two
conditions coincide (Fig.~\ref{fig:gates}). First, the cap interrupts an action before the agent
completes it. Second, the persistent runtime allows the matched agent to continue from partial state,
while the stateless runtime causes the mismatched agent to restart. We call these conditions
\emph{truncation exposure} and \emph{continuation advantage}. They diagnose the cap amplification we
measure, not every quality difference between runtimes: Rule Diagnosis carries a substantial runtime
gap at both caps while never reaching the cap at all
(App.~Table~\ref{tab:crosstaskq}). We treat the second condition as an empirical diagnostic supported
by four designed settings, not as a general theorem.

\textbf{Condition 1: does the cap interrupt the agent's actions?} We estimate truncation exposure by
replaying each action block from the slack-cap ($c{=}80$) runs and counting how
many tool calls it would make (App.~\ref{app:replay}). The exposure $E_c$ is the fraction of those
blocks that a cap of $c$ would split. $E_{25}$ is
$0.53$ for Opaque Knapsack,
$0.04$ for Navigation and
$0.00$ for Rule Diagnosis.

The task's total tool budget does not determine exposure. Rule Diagnosis permits about
$30$ probes against a cap of $25$, but the agent
divides them among short probe-and-check cycles of its own accord. Its widest recorded action block is
$19$ calls, and it never reaches the per-turn cap, in either
condition, in any episode. The Knapsack agent instead writes the whole inspection sweep as one long
loop. The cap therefore acts on the number of calls in each action block, not on the total number of
calls in the episode.

All three $P\!\to\!S$ conditions repeat work, and two of the three pay for it in tokens at both caps,
but only Opaque Knapsack reaches the cap often enough for that redundancy to become a quality collapse.
Rule Diagnosis is the exception on cost: its mismatched episodes are shorter at the binding cap and
spend fewer tokens there than matched ones (App.~Table~\ref{tab:crosstaskq}). Navigation's exposure is
low but not zero, so the cap does occasionally interrupt an action, just too rarely to produce a
detectable outcome response. The Navigation and
Rule Diagnosis controls use $n{=}16$ tasks, one seed and one rollout, so we report
that we did not detect an effect rather than claiming that none exists (App.~\ref{app:diag} defines
$E_c$ and gives its pooling and right-censoring caveats).

\textbf{Condition 2: does persistent state allow the agent to continue?} Exposure alone is not
enough. The matched Knapsack condition has appreciable exposure too
($E_{25}=0.32$). Yet matched quality
falls only modestly, from $0.77$ to
$0.66$, because under a persistent runtime the partial working record
is still in interpreter variables on the next turn and the agent resumes instead of restarting.
Narrowing an interface makes the same point: we rebuilt the same $16$ Navigation instances with the topology channel's batch
limit cut to $2$ nodes per \texttt{neighbors} call, a value fixed before any new run
(App.~\ref{app:crossfam}). Exposure appears as predicted and both runtimes collapse. The intervention
moves cap amplification by $+0.25$
$[-0.18,+0.70]$, an interval spanning
zero, so we do not detect mismatch-specific amplification.

\textbf{Post-cap behavior distinguishes the settings.} We count a restart when the turn after a cap
hit reissues the task's listing call before acquiring anything new (App.~Table~\ref{tab:mech}). On
Knapsack the restart rate is $0.84$ under
$P\!\to\!S$ and $0.00$ under $P\!\to\!P$, a difference
of $0.84$
$[0.70,0.95]$.
With batch-limited Navigation both conditions restart
($0.94$ against
$0.98$, a difference of
$-0.04$
$[-0.14,+0.02]$,
App.~Table~\ref{tab:crosstask}). The matched condition still names the surviving variables in its next
block on $100\%$ of those turns. Both runtimes deteriorate
under frequent truncation there, and we do not detect additional mismatch-specific amplification.
Re-derivation is one operational diagnostic of
continuation advantage, not its definition (App.~\ref{app:diag}). Knapsack is the only setting that
meets both conditions.

\textbf{Intervention: is preserving the variables enough?} The post-cap continuation evidence above is
observational, so we next intervene directly on the runtime boundary. A third runtime, \emph{silent
carryover}, resets every turn except one the cap truncated, whose interpreter variables survive into
the next turn only, with the prompt, the observation format and the reset-state banner unchanged
(App.~\ref{app:ckpt}).

Preserving the variables removes the missing-variable errors the reset causes, and the turns
raising one after a truncation fall from
$6.2$ per episode to
$0.0$. It recovers
only $17\%$
$[1\%,36\%]$ of the quality gap, and
it does not reduce restarting. Telling the agent which variables survived lowers the restart rate from
$0.86$ to
$0.31$ of post-cap turns, but final quality
does not move ($-0.001$
$[-0.129,+0.121]$). Under the
tested policy and reset-oriented prompt, preserving or naming the surviving variables was not enough to
restore matched performance, which motivates learning an explicit policy for continuing the interrupted
computation.

Each intervention is one rollout over $n{=}25$ tasks, and the
$17\%$ is descriptive gap closure rather than a causal mediation estimate.
App.~\ref{app:ckpt} reports the design, the pre-specification, both baselines, the intervals and the
run-to-run variation.

Needing a working record is not enough: the evaluation must also reach a point where the matched and
the mismatched agent differ in whether they can continue. Both the width of an action block and the
behavior after an interruption are properties of the agent--task pair, to be measured rather than read
off a nominal budget.

\paragraph{Implications for evaluation.}\label{sec:protocol}
These results imply that an interactive-agent score should be read as conditional on both the runtime
contract and the resource limits. The configured cap alone does not establish whether anything is interrupted. What
matters is whether the agent's action blocks actually cross a boundary at which the runtime contracts
differ. Evaluations should state the persistence semantics, per-turn cap, episode horizon and
termination conditions, and report task quality alongside tokens, calls and repeated work. Where
runtime dependence is a plausible confound, varying the cap or cross-evaluating runtimes separates a
robust policy from one whose success depends on a particular execution contract.

\section{Limitations}
\label{sec:limitations}
\textbf{What the design isolates.} The within-adapter cap sweep holds the trained model and the
deployment runtime fixed, so it isolates the effect of the cap. Reading that response as
\emph{mismatch-specific} still rests on the matched-runtime and stateless-trained controls.
Cross-runtime comparisons do not separate interpreter persistence from the runtime-specific prompt
package that announces it. Comparisons between persistent- and stateless-trained adapters do not isolate persistence alone, because the two
training conditions also differ in instructions, few-shot demonstrations, trace length and asymmetric
training-trace truncation (stateless traces run $\approx2.3\times$ longer, and
$117$ of $1{,}000$ carry a truncation signature against none of the persistent ones). Separating them needs length-matched
training and prompt-controlled runtime comparisons. The two cap settings were fixed in
advance from task mechanics but were not preregistered, so we report the continuous $\log c$ trend
alongside them (Fig.~\ref{fig:dose}, Sec.~\ref{sec:design}).

\textbf{Scope of the evidence.} The primary factorial uses one procedurally generated task family and
one LoRA recipe. The Mistral and Llama replications are single-seed and single-rollout at
$n{=}25$, and differ from the main replication in ways their base checkpoints
forced (App.~\ref{app:arms}). The additional-task controls at $n{=}16$ and the
carryover interventions run only on the main replication
(Apps.~\ref{app:crossfam}--\ref{app:ckpt}). Every adapter is also trained on traces generated by one
teacher model and one prompting recipe, so a different teacher could produce different action widths or
cap sensitivity.

\textbf{Untested runtime conditions.} Results may also depend on how the runtime interrupts an action
and on the finite turn, wall-clock and context budgets. Our runtime aborts a Python block at a
tool-call boundary and returns the partial output. We do not test automatic continuation or explicit
checkpointing, and carryover restores the variables a truncation wiped, not the interrupted
computation (App.~\ref{app:ckpt}). Exact effect sizes are therefore likely task- and harness-specific. What we mean to generalize is the
evaluation question, not the magnitudes observed here.

\section{Conclusion}\label{sec:conclusion}
Agent evaluations usually hold the harness fixed while varying the model. For
agents trained on interaction traces, that separation is incomplete: the runtime contract becomes part of
the learned interface. In our evaluation a persistent-trained agent on the stateless runtime
scored $0.61$ of optimal under a slack cap and
$0.07$ under a binding one, against
$0.77$ and $0.66$ matched. Trace replay showed the mechanism: the cap forced an intended computation across a turn
boundary, the reset erased the partial working record, and the agent rebuilt information it had already
acquired instead of resuming. This cap amplification reproduced across training seeds, an independent
second rollout and two further base models. Across the tasks and controls studied, cap tightening
amplified mismatch only when the cap truncated an intended action \emph{and} matched execution resumed
from surviving state whereas mismatched execution restarted.

Fixing one configuration can therefore conceal whether an agent is robust to runtime transfer or
merely compatible with one harness. The results also motivate a training target: agents should learn
not only to exploit persistent state but to checkpoint and resume when persistence is uncertain. An
interactive-agent score belongs to the agent and to the harness it runs in, and runtime-transfer tests
reveal how strongly measured performance depends on
compatibility with the execution environment.

\bibliographystyle{plainnat}
\bibliography{references}

\appendix
Appendices~\ref{app:tasks}--\ref{app:train} specify the tasks, runtimes, training data and inference
setup. App.~\ref{app:results} reports the complete primary results and the statistical procedures.
Appendices~\ref{app:mech}--\ref{app:controls} give the trace analysis, the cross-task controls and
the continuation interventions supporting Sec.~\ref{sec:mech}. App.~\ref{app:robust} reports robustness
and replication, App.~\ref{app:repro} documents the released artifacts,
App.~\ref{app:impact} states the broader impact, and App.~\ref{app:notation} collects the notation and
terminology used throughout.

\section{Tasks and runtime contracts}
\label{app:tasks}

\subsection{What makes a task diagnostic, and how instances avoid collapse}
\label{app:task_design}

A task is not CodeAct-diagnostic merely because it exposes an API. If the optimal interaction is a fixed
sequence of tool calls followed by one static computation, the task measures program synthesis and tool
syntax but not stateful reasoning between observations, the distinction multi-turn tool-interaction
benchmarks \citep{yao2024taubenchbenchmarktoolagentuserinteraction,wang2024mintevaluatingllmsmultiturn} are built to test. Each of our
three families is therefore built so that tool feedback must be accumulated into a structured working
record, later actions depend on that record, and the score distinguishes rebuilt work from new
progress.
The task constraints and the terminal verifier are identical under both runtimes, so persistence is
never baked into the ground truth. Retained training instances additionally have a filter-passing
teacher trace under each contract (App.~\ref{app:train}). Table~\ref{tab:taskspec} states, per family, what is latent, what
the working record holds, what the verifier checks, and the structural constraint that keeps instances
from being trivially solvable. Every family uses seeded, released instances and a terminal verifier
scoring against ground truth.

\begin{table}[h]
\centering\small
\caption{The three families as working-record tasks. The anti-collapse constraint is enforced per
instance by construction, except for Navigation, where it is \emph{best-effort}: if the generator
exhausts its candidate pool it returns the closest candidate with a recorded warning, so a minority of
Navigation instances may be easier than targeted.}
\label{tab:taskspec}
\begin{tabular}{p{0.13\columnwidth} p{0.19\columnwidth} p{0.17\columnwidth} p{0.17\columnwidth} p{0.22\columnwidth}}
\toprule
Task & Latent information & Working record & Verifier & Key anti-collapse constraint \\
\midrule
Opaque Knapsack & Item attributes and the allowed-class set & Inspected-item table & Exact DP optimum over allowed items & Non-degenerate optimum: optimal set $\ge3$ items, no item above $40\%$ of optimal value \\
Navigation & Graph mechanics: traps, keys, locks & Map, keys, traps, locks & Goal / shortest-path progress score & Exactly $1$ locked gate into the goal, its key reachable trap-free \\
Rule Diagnosis & Hidden piecewise rule & Probe map and hypothesis & Exact latent rule & $\ge3$ intervals of length $\ge6$, at least two internal breakpoints \\
\bottomrule
\end{tabular}
\end{table}

\subsection{Opaque Knapsack: objective, tool API, and evaluation distribution}
\label{app:task}

\paragraph{Formal task.} An Opaque Knapsack instance is a tuple $(\mathcal{I},C,\mathcal{K},\mathcal{A},B)$:
a set of items $\mathcal{I}=\{i_1,\dots,i_n\}$ with opaque IDs, a capacity $C\in\mathbb{R}^+$, a class
universe $\mathcal{K}$, a \emph{hidden} allowed-class subset $\mathcal{A}\subseteq\mathcal{K}$, and an
inspection budget $B\in\mathbb{N}$. Each item $i$ has latent attributes $(w_i,v_i,c_i)$ (weight, value,
class) revealed only by inspecting it. The agent selects $S\subseteq\mathcal{Q}$ (where $\mathcal{Q}$ is
the set of inspected items) maximizing $\sum_{i\in S}v_i$ subject to $\sum_{i\in S}w_i\le C$,
$c_i\in\mathcal{A}\;\forall i\in S$, and $|\mathcal{Q}|\le B$. The score is normalized optimality
$\sum_{i\in S}v_i / \mathrm{OPT}$ (a 0/1-DP reference restricted to allowed-class items), clamped to $0$
on any capacity overflow. An instance is \emph{solved} iff the optimality gap is $0$.

\paragraph{Initial observation.} An episode opens with a single task message stating three instance
quantities and nothing else: the capacity $C$, the inspection budget $B$, and the class alphabet, listed as
the class symbols realized among that instance's items (a subset of the $26$ letters, since a class is drawn
per item). The message does not list item IDs, which the agent retrieves with the free
\texttt{list\_items()} call. The allowed set $\mathcal{A}$ is genuinely hidden and is displayed by no tool: \texttt{inspect}
returns $(w_i,v_i,c_i)$ but not whether $c_i\in\mathcal{A}$, so the only admissibility signal is the
outcome of a \texttt{take\_item} attempt, and that signal is asymmetric. A disallowed class raises a
named error (\texttt{"Item class 'X' not allowed."}) that is checked before any state change, so a refused
take leaves the selection, the running weight and the score untouched and costs only the call and the turn
it occupies. An allowed class instead succeeds, which both confirms admissibility and adds the item
irreversibly, since the API has no drop operation: a probe that lands on an allowed class commits that
item's weight against $C$, and probing with a poor item can put the optimum out of reach. Probing also
identifies $\mathcal{A}$ only up to the classes realized in the instance, as allowed classes that no item
happens to carry appear neither in the opening alphabet nor in any observation. They are equally
irrelevant to the objective, which ranges over existing items alone. These class verdicts, like the
inspected attributes, are part of what the working record must carry across a reset. Because
\texttt{take\_item} likewise refuses any take that would exceed $C$, the overflow clamp in the score is a
defensive guard rather than a reachable failure mode.

\paragraph{Tool API.} Agents act exclusively through three task tools plus the runtime-provided
\texttt{finish()}. Table~\ref{tab:knaptools} gives the contract. The load-bearing property for our
mechanism (Sec.~\ref{sec:mech}) is that \texttt{inspect} is \emph{budgeted but cached}: the budget counts
\emph{distinct} items, and re-inspecting an already-seen ID returns the cached record and consumes
neither budget nor a fresh environment query. This is what makes textual reconstruction of a wiped
working record an \emph{idempotent} channel rather than a fresh-information channel.

\begin{table}[h]
\centering\small
\caption{Opaque Knapsack tool API. Tool return values are available to the executing Python block. The
next observation carries the captured \texttt{stdout} and any uncaught execution error, so a return
value is not added to the transcript automatically and must be printed to stay visible after a reset.
Every tool call counts against the per-turn cap, including the calls below that spend no inspection
budget. Budget $B$ counts distinct inspected IDs.}
\label{tab:knaptools}
\begin{tabular}{p{0.34\columnwidth} p{0.58\columnwidth}}
\toprule
\textbf{Call $\rightarrow$ return} & \textbf{Budget / constraints} \\
\midrule
\texttt{list\_items() $\rightarrow$} JSON array of all opaque item IDs & Free (no budget cost). \\
\texttt{inspect(id) $\rightarrow$} JSON \texttt{\{"weight","value","class"\}} & Consumes one unit of $B$ on the \emph{first} inspection of an ID, and repeat inspections are cached and free. Errors if $B$ is exhausted or the ID is unknown. \\
\texttt{take\_item(id) $\rightarrow$} None & Adds the item. Errors if already taken, of a disallowed class, or if it would exceed $C$. Every rejection is checked before any state change, and there is no drop tool, so an accepted take is irreversible. \\
\texttt{finish() $\rightarrow$} None & Terminates the episode (runtime-provided). \\
\bottomrule
\end{tabular}
\end{table}

\paragraph{Evaluation distribution.} Each instance samples: (i) $n\in[40,65]$ items; (ii)
weights $w_i\sim\mathrm{U}\{5,\dots,35\}$; (iii) values $v_i\sim\mathrm{U}\{10,\dots,300\}$; (iv) a class
per item drawn uniformly from the $26$-symbol alphabet $\{A,\dots,Z\}$, with $|\mathcal{A}|=4$ allowed
classes; and (v) capacity $C=\mathrm{round}\big(\rho\sum_{i:c_i\in\mathcal{A}} w_i\big)$ with
$\rho\sim\mathrm{U}[0.35,0.5]$ over the total weight of allowed-class items only (floored at the minimum
allowed item weight). The inspection budget is derived per instance from the capacity, the mean weight
and the allowed-class fraction. Under the evaluated distribution it commonly saturates at the number of
items, so solving the task typically requires inspecting most of the catalog and, at a binding
per-turn cap, reconstruction must span several turns: the condition under which statelessness costs
quality.
The closed form and the exact per-instance budgets are in the released task definitions.

\subsection{Runtime contracts and evaluation prompts}
\label{app:runtime}

\paragraph{Execution loop.} For a model call that returns an executable response, the interaction is
Reflection (free text) $\to$ Action (one fenced Python block) $\to$ Observation. The agent extracts fenced code blocks and executes \emph{only the first}. Additional
blocks are discarded and flagged in the observation, and a turn with no block yields a structured
retry-prompt. Episodes terminate when the model calls \texttt{finish()} inside a block, or on reaching
the maximum turn count $T_{\max}$. Interpreter exceptions (e.g.\ \texttt{NameError}) are returned as
structured observations, not fatal errors, so the model can repair on the next turn. Every observation
carries a JSON header
\begin{lstlisting}
{"observation": {"success": bool, "result": ..., "output": "...", "error": ...},
 "runtime_state": {"runtime": "persistent" | "reset",
                   "active_globals": [...], "last_step_globals": [...]}}
\end{lstlisting}
\texttt{last\_step\_globals} lists the variable \emph{names} that existed after the previous block in
\emph{both} runtimes, while \texttt{active\_globals} lists names currently executable in the interpreter. Only
names, never values, are exposed, so both runtimes receive identical structural metadata about the prior
step. The interpreter enforces persistence or reset independently of the prompt text.

\paragraph{What differs between the conditions.} The system context is assembled, in order, from four
messages: a shared base prompt, a runtime-specific banner carrying the state rule, a runtime-specific
few-shot block, and a tool prompt carrying the per-turn cap line. The task goal is identical across
conditions and delivered as the first user message. Table~\ref{tab:promptdiff} is the audit of that
bundle. One representative pair of few-shot blocks:
\begin{lstlisting}
--- persistent (Turn 2) ---   active_globals=["items","foo"] observed
"I can see items and foo in active_globals, so I reuse them directly."
items = foo(items); total = sum(items); print(f"Total: {total}"); finish()

--- stateless (Turn 1 -> 2) --- active_globals=[] after reset
Turn 1: print(f"STATE: items={items}")   # print state to survive the reset
Turn 2: "active_globals is empty. I reconstruct items and foo from the previous observation."
        items = [10, 20]; ... ; finish()
\end{lstlisting}
The full base prompt, both banners, both few-shot blocks and the tool prompt are released verbatim
with the code, dumped from the running agent rather than transcribed.

\begin{table}[h]
\centering\small
\caption{What the two conditions change, and what the cap changes. Within a runtime condition, changing
$c$ changes its announced value and its enforcement, and every other prompt component and resource
limit stays fixed (App.~\ref{app:cap}). The persistent-vs-stateless comparison is different: it
\emph{bundles} the top four rows with the interpreter's actual reset behavior, which is why we read
every cross-adapter quantity as bundled and say so in the Limitations.}
\label{tab:promptdiff}
\begin{tabular}{p{0.21\columnwidth} p{0.26\columnwidth} p{0.26\columnwidth} c}
\toprule
Component & Persistent condition & Stateless condition & Changes with cap? \\
\midrule
Interpreter bindings & Survive the turn boundary & Cleared each turn & No \\
\texttt{active\_globals} & Prior names shown as active & Empty & No \\
State rule & ``globals persist; never re-paste code'' & ``state resets; redefine each step'' & No \\
Few-shot behavior & Reuses variables across turns & Prints and rebuilds state & No \\
Tool-cap instruction & Same wording with $N$ substituted & Same & \textbf{Yes} \\
\bottomrule
\end{tabular}
\end{table}

\section{Training and evaluation details}
\label{app:train}

\paragraph{Teacher traces and pairing.} Training traces come from a Gemini 3 Flash \citep{gemini3flash2025}
teacher (\texttt{gemini/gemini-3-flash-preview}) run inside the same CodeAct harness, persistence
contract, base prompt, mode banner and runtime-matched few-shot demonstrations used at evaluation
(App.~\ref{app:runtime}): one rollout per (task instance, training runtime). Generation cached the
teacher's raw responses locally, but that cache is not part of the release (App.~\ref{app:repro}): what we
release is the processed, instance-paired SFT data itself, and that fixes the training data exactly.
Teacher decoding was unseeded, so the original generation is not independently reproducible. The
teacher generated at a per-turn cap of $25$, which is the binding deployment cap
($c{=}25$), so the binding train$\to$deploy comparison introduces no cap
shift and only the runtime condition changes. At $c{=}80$, and at every
dense-sweep cap other than $25$, deployment uses a per-turn capacity absent from
teacher generation. The within-condition cap response remains unaffected because it holds the adapter
fixed. Training instances are disjoint from the held-out
evaluation pool: generation runs before and independently of evaluation, from a separate generator-seed
range.

The key design choice is \emph{instance pairing}: we intersect persistent and stateless traces by task ID
and retain a pair only if \emph{both} sides independently pass the filters, yielding
instance-matched training data rather than merely runtime-matched pools. The retained subset is fixed by
a seeded shuffle of the sorted task-ID intersection, so the training set is deterministic and
regeneratable. Table~\ref{tab:traindata} gives the yield.

\begin{table}[h]
\centering\small
\caption{Training-data construction (the candidate and dual-valid counts are Opaque Knapsack's, and the
two control families have their own paired pools, retained at the same size by the same recipe and
evaluated only on their own family). A trace is
valid only if it scores $\ge0.5$, calls \texttt{finish()} within its
last three steps, contains no near-duplicate recovery loop (character match ratio
$>0.9$ against any of the previous four assistant messages, from Python's
standard-library sequence matcher at its default settings), and has
bad-error density $\le0.1$, counting interpreter errors but
\emph{excluding} tool-contract errors (an exhausted inspection budget, say) and
``\texttt{Tool call limit exceeded}'', which are legitimate feedback.}
\label{tab:traindata}
\begin{tabular}{ll}
\toprule
Setting & Value \\
\midrule
Teacher & Gemini 3 Flash (\texttt{gemini/gemini-3-flash-preview}) \\
Teacher runtime & Runtime-matched (one rollout per instance per runtime) \\
Teacher cap / turns (Opaque Knapsack) & $25$ / $40$ \\
Candidate pairs & 1{,}900 \\
Dual-valid pairs & 1{,}109 (58.4\% paired yield) \\
Retained pairs & 1{,}000 per training runtime (seeded shuffle, seed $42$) \\
Preparation length & 16{,}384 tokens (head--tail truncation, assistant targets never masked) \\
Mean trace length & Persistent $4.6$k, stateless $10.7$k tokens \\
Mean assistant turns & Persistent $4.3$, stateless $6.1$ \\
\bottomrule
\end{tabular}
\end{table}

That trace-length asymmetry is the most important qualification of the cross-adapter comparison.
Measured on the released Opaque Knapsack traces with the Qwen3-8B tokenizer, accepted
stateless traces are $\approx\!2.3\times$ longer than persistent ones,
because stateless-runtime traces carry explicit state re-emission. The stateless tail reaches
$16.2$k tokens and persistent traces reach at most
$12.9$k, against a preparation window of exactly
$16{,}384$ tokens less a $100$-token safety margin, so preparation
truncates a trace above $16{,}284$. That ceiling binds on one side only. In the
released Opaque Knapsack training sets, $117$ of the $1{,}000$ stateless traces carry a truncation
signature, either a masked observation or a dropped middle turn, and none of the $1{,}000$ persistent
traces carry either. The persistent and stateless
adapters therefore differ in training-trace length as well as in runtime semantics, and we read every
cross-adapter quantity as bundled.

\begin{table}[h]
\centering\small
\caption{Fine-tuning and inference. Settings are shared across all replications unless a column says
otherwise. The two base-model replications retrain the same recipe on a different base checkpoint
(App.~\ref{app:arms}). The
compute totals cover different numbers of adapters: 6 for
Qwen3-8B and two for each base-model replication, with the
4 control-task adapters adding a further
$\approx10.71$ GPU-h outside this table. All arms are served
with vLLM \citep{kwon2023efficientmemorymanagementlarge}. Serving compute was
never metered and is not reported.}
\label{tab:trainconfig}
\begin{tabular}{lccc}
\toprule
& Qwen3-8B & Mistral-7B-v0.3 & Llama-3.1-8B \\
\midrule
Base checkpoint & \texttt{Qwen/Qwen3-8B} & \texttt{mistralai/} & \texttt{meta-llama/} \\
 & & \texttt{Mistral-7B-v0.3} & \texttt{Llama-3.1-8B} \\
Adaptation & \multicolumn{3}{c}{QLoRA, 4-bit NF4, double quantization, bfloat16 compute} \\
Rank / $\alpha$ / dropout & \multicolumn{3}{c}{$r{=}64$ / $\alpha{=}128$ / $0.05$} \\
Modules & \multicolumn{3}{c}{All attention and MLP projections ($q,k,v,o,\text{gate},\text{up},\text{down}$)} \\
Epochs / optimizer / LR & \multicolumn{3}{c}{$3$ / AdamW, cosine, peak $1\times10^{-4}$, warmup $0.03$, decay $0.01$} \\
Micro-batch / accumulation & \multicolumn{3}{c}{$1$ / $16$} \\
Sequence length & 16{,}384 & 16{,}384 & 16{,}384 \\
Serving window & 40{,}960 & $32{,}768$ & 40{,}960 \\
Serving (vLLM, bfloat16) & \multicolumn{3}{c}{Temperature $0.2$, $8{,}192$ output tokens/step, $T_{\max}{=}40$, $600$\,s per call} \\
Episode wall-clock & \multicolumn{3}{c}{$3{,}000$\,s, raised to $4{,}000$\,s for the $T_{\max}{\in}\{80,160\}$ controls} \\
Training compute & $\approx20$ GPU-h & $\approx7.94$ GPU-h & $\approx8.57$ GPU-h \\
\bottomrule
\end{tabular}
\end{table}

We fine-tune Qwen3-8B \citep{yang2025qwen3technicalreport} with QLoRA
\citep{dettmers2023qloraefficientfinetuningquantized} and LoRA
\citep{hu2021loralowrankadaptationlarge} on $4\times$NVIDIA
GH200 GPUs (Table~\ref{tab:trainconfig}). Persistent- and stateless-trained adapters differ only in
their
training data. Each task family has its own persistent/stateless pair, trained on that family's paired
traces and evaluated only on it. Across the paper that is 14 adapters.
Training the 4 cross-family (Navigation and Rule Diagnosis)
adapters took $\approx10.71$ GPU-h total on
$4\times$GH200, computed as each run's logged
elapsed wall-clock multiplied by the 4 GPUs.
The three seeds reported as seed-0/777/1337 are training-config seeds $3407$,
$777$ and $1337$, which share data and hyperparameters and
differ only in the seed. All inference settings are identical across the four train$\to$runtime
conditions and across caps. Across conditions, only the adapter weights, the runtime condition
(interpreter semantics together with its prompt package) and the cap configuration vary.

\section{Complete primary results and statistical procedures}
\label{app:results}

\subsection{The \texorpdfstring{$2\times2$}{2x2} experiment, with outcome and operational metrics}
\label{app:main2x2}

Table~\ref{tab:main2x2} reports all four conditions at both caps. It gives the per-condition
intervals, the operational counters and the termination taxonomy that Table~\ref{tab:main} in the main
text summarizes into two rows.

\begin{table}[h]
\centering\small
\caption{\textbf{The $2\times2$ experiment at a binding ($c{=}25$) and slack
($c{=}80$) cap} (Opaque Knapsack). Quality is normalized optimality on the main
evaluation (seed-0, $n{=}25$ paired instances) with task-paired bootstrap
$95\%$ CIs (resampling task instances, from a single
temperature-$0.2$ rollout per condition with no sampling seed, not decoding
variability). Operational columns are per-episode means over the same instances. \emph{Steps} counts
logged steps, which are the model calls that returned a usable completion, not the model calls the
episode attempted: a call that fails or is rejected spends a turn of the horizon without logging a step
(App.~Table~\ref{tab:horizon}). \emph{Listing calls} counts \texttt{list\_items()} occurrences in the
executed code, including the opening listing every episode makes, so a condition that never
re-reconstructs reads $1.0$. \emph{Cap-hit} is the mean per-episode fraction of logged steps reaching
the cap and
\texttt{NameErr} counts steps raising a missing-variable error. \emph{Tokens} sums input and output
tokens across accepted model calls, including transcript tokens processed again on later calls, which
dominate the total: input is $95$--$97\%$ of it in every cell here. Termination is finish/turn-limit/timeout counts, and timed-out episodes retain
their partial score rather than zero. The collapse condition
($P\!\to\!S$ @ $25$) records $30.8$ listing calls per episode
yet inspects fewer distinct items, spends the most tokens, and terminates by turn-limit. The
binding-cap interaction replicates across three training seeds (Sec.~\ref{sec:main}).}
\label{tab:main2x2}
\setlength{\tabcolsep}{3.5pt}
\begin{tabular}{llccccccc}
\toprule
Cap & Condition & Quality (95\% CI) & Steps & Listing calls & Tokens & Cap-hit\% & \texttt{NameErr} & Fin/turn/to \\
\midrule
\multirow{4}{*}{$c{=}25$}
 & $P\!\to\!P$ (matched)  & $0.66\ [0.55,0.77]$          & 9.8  & 1.1           & 113k & 41 & 0.04 & 19/6/0 \\
 & $P\!\to\!S$ (mismatch) & $\mathbf{0.07}\ [0.01,0.16]$ & 35.2 & \textbf{30.8} & 498k & 71 & \textbf{5.44} & 2/23/0 \\
 & $S\!\to\!P$            & $0.12\ [0.03,0.22]$          & 18.6 & 1.0           & 288k & 36 & 0.00 & 6/16/3 \\
 & $S\!\to\!S$ (matched)  & $0.26\ [0.17,0.35]$          & 11.2 & 1.0           & 172k & 15 & 0.00 & 6/10/9 \\
\midrule
\multirow{4}{*}{$c{=}80$}
 & $P\!\to\!P$            & $0.77\ [0.66,0.86]$          & 4.6  & 1.0           & 37k & 0 & 0.00 & 24/1/0 \\
 & $P\!\to\!S$            & $\mathbf{0.61}\ [0.45,0.77]$ & 13.0 & 7.9           & 162k & 4 & 3.80 & 18/7/0 \\
 & $S\!\to\!P$            & $0.23\ [0.13,0.35]$          & 10.0 & 1.0           & 128k & 0 & 0.00 & 10/6/9 \\
 & $S\!\to\!S$            & $0.51\ [0.36,0.65]$          & 9.8  & 1.0           & 146k & 0 & 0.00 & 12/8/5 \\
\bottomrule
\end{tabular}
\end{table}

\subsection{Trace-derived quantities}
\label{app:diag}

The operational counters are derived by textual matching on the trace. Table~\ref{tab:diag} gives their
definitions and aggregation units. The literal match strings and the parsing implementation are documented with
the analysis code.

\begin{table}[h]
\centering\small
\caption{Trace-derived quantities.}
\label{tab:diag}
\begin{tabular}{p{0.26\columnwidth} p{0.66\columnwidth}}
\toprule
Quantity & Definition \\
\midrule
Cap hit & The logged step's interpreter error reports the per-turn tool-call limit exceeded, and a cap-rejected call does \emph{not} decrement the task inspection budget (App.~\ref{app:cap}) \\
Missing binding (\texttt{NameErr}) & The logged step's interpreter error contains \texttt{NameError}, the signature of a reset destroying a binding the policy assumed persisted \\
Post-cap restart & The next executed turn after a cap hit reissues the task's index or listing operation (\texttt{list\_items}, \texttt{neighbors}) before acquiring anything new. This is conditional on a cap hit, unlike the unconditional listing-call count in Table~\ref{tab:main2x2} \\
Termination & Exactly one of: \texttt{finish()} called, maximum turns reached, or episode wall-clock timeout \\
Exposure $E_c$ & The fraction of a condition's slack-cap action blocks whose replayed call count exceeds $c$ (App.~\ref{app:replay}) \\
\bottomrule
\end{tabular}
\end{table}

Two readings need care. A timed-out episode is scored by the environment on its final state, so it
\emph{retains its partial score} rather than being zeroed, and it is not retried. Only its terminal
label changes. And $E_c$ is \emph{pooled over turns, not episodes}, so conditions whose episodes run longer
contribute more blocks (on Opaque Knapsack $P\!\to\!S$, $E_{25}$ is
$0.53$ pooled against
$0.66$ per episode). The replayed count is also
\emph{right-censored} by the slack cap itself, affecting
$4\%$ of Opaque Knapsack $P\!\to\!S$ blocks and none
elsewhere. Censoring leaves $E_{25}$ unaffected, since a censored block exceeded
$25$ either way, but it makes the reported mean and maximum block widths lower
bounds.

\subsection{Estimators, resampling units, and per-seed results}
\label{app:stats}

Every quantity is paired over the common task IDs of the four conditions at a given seed and cap. We report
the persistent model's runtime gap $G_P{=}Q_{PP}{-}Q_{PS}$, the train$\times$runtime interaction
$D{=}(Q_{PP}{-}Q_{PS}){-}(Q_{SP}{-}Q_{SS})$, and its moderation
$T{=}D(c_{\text{bind}}){-}D(c_{\text{slack}})$. Because the three seeds were evaluated on the
\emph{same} task IDs, seed and task are crossed, not nested: the pooled $T$ interval resamples seeds and
task IDs independently and retains the selected seed$\times$task scores (a crossed bootstrap,
$5{,}000$ resamples). Every other resampling unit is a task instance.
Per-condition and per-cap intervals are seed-0 task-only bootstraps and carry no training-seed variation.
Evaluation is a \emph{single, unseeded} temperature-$0.2$ rollout per (task,
condition). The bootstrap resamples those observed outcomes, which already carry decoding randomness,
so no interval here separately estimates within-task decoding variability. App.~\ref{app:arms}
assesses sensitivity to it. With only three seed levels we treat the per-seed \emph{sign} replication as the primary
multi-seed evidence and the crossed-bootstrap interval as secondary.

\paragraph{The common set, and how it was selected.} We report seed-0 at its full
$n{=}25$ task set and the three seeds together only on a common task set
(Table~\ref{tab:seedcells_rep}). That set is the first $8$ task
IDs (generation indices $0$--$7$), the prefix on which every cell of the
factorial was run. The extra seeds were run deeper in most persistent-runtime cells, on a
$20$-instance prefix, but in their stateless-runtime cells only the first
$8$ instances were ever dispatched. The remainder were never launched, so they are
missing rather than failed, and no episode that ran is dropped: timed-out and turn-limited episodes keep
their partial scores here as everywhere else in the paper. Because
instances are generated before and independently of evaluation, this is a \emph{pre-specified} prefix,
not the subset of episodes that happened to finish. The selection is outcome-independent and cannot
introduce completion-driven bias. On the main evaluation, $D$ falls from
$+0.734$ at the binding cap to
$+0.426$ at the slack cap, giving
$T{=}+0.309$
$[+0.041,+0.567]$. On the common set $T$
is positive in every seed
($+0.368,+0.474,+0.512$).
The crossed seed$\times$task bootstrap gives $T{=}+0.45$
$[+0.13,+0.71]$, but with three seeds and
$8$ tasks we do not rely on that interval. Seed-0 is the higher-powered estimate and
is never pooled with the common set.

\begin{table}[h]
\centering\small
\caption{Per-seed condition means (normalized optimality) and the interaction $D$. The first row is the
primary seed-0 evaluation at $n{=}25$. The rest are all three seeds on the \emph{same}
common $8$ task IDs (generation indices $0$--$7$,
pre-specified and outcome-independent). Seed-0's common-set conditions differ from its primary row because that
is the smaller sample. Per-seed $T{=}D(25){-}D(80)$ is
positive in every seed.}
\label{tab:seedcells_rep}
\begin{tabular}{llccccc}
\toprule
Seed & Cap & $Q_{PP}$ & $Q_{PS}$ & $Q_{SP}$ & $Q_{SS}$ & $D$ \\
\midrule
\multicolumn{7}{l}{\emph{Primary evaluation, seed-0, $n{=}25$}}\\
0 & 25 & 0.662 & 0.075 & 0.115 & 0.262 & $+0.734$ \\
0 & 80 & 0.766 & 0.615 & 0.232 & 0.505 & $+0.426$ \\
\midrule
\multicolumn{7}{l}{\emph{Sign-replication set, common $8$ task IDs}}\\
0    & 25 & 0.722 & 0.039 & 0.166 & 0.109 & $+0.625$ \\
0    & 80 & 0.796 & 0.752 & 0.112 & 0.326 & $+0.257$ \\
777  & 25 & 0.741 & 0.041 & 0.134 & 0.147 & $+0.713$ \\
777  & 80 & 0.704 & 0.665 & 0.144 & 0.343 & $+0.239$ \\
1337 & 25 & 0.757 & 0.041 & 0.166 & 0.200 & $+0.750$ \\
1337 & 80 & 0.757 & 0.699 & 0.187 & 0.367 & $+0.238$ \\
\bottomrule
\end{tabular}
\end{table}

\subsection{Dense cap sweep}
\label{app:dense}

Table~\ref{tab:dense} gives the numbers behind Fig.~\ref{fig:dose}. The same
$12$ task IDs are evaluated in \emph{both} conditions at every cap (fully paired across
the sweep). The raw means rise but are not strictly monotone, which is why we summarize by a continuous
regression rather than any single pairwise comparison: we fit the \emph{task-panel} model
$Q_{ic}=\alpha_i+\beta\log c+\epsilon_{ic}$ with a fixed effect $\alpha_i$ per task ID over all
$12\times8$ task--cap observations, giving
$\beta{=}+0.299$
$[+0.229,+0.371]$. The interval comes from a bootstrap
that resamples \emph{task IDs} and retains each sampled task's full eight-cap vector, preserving the
across-cap pairing, and refits the slope on each resample. We do not resample scores independently
within a cap, which would break that pairing. The matched $P\!\to\!P$ curve, fit the same way on the
same task IDs, is comparatively flat ($\beta{=}+0.095$,
$[-0.011,+0.189]$), and the paired slope
difference is $\beta_{PS}-\beta_{PP}{=}+0.204$
$[+0.094,+0.309]$, excluding $0$: the
interaction is visible across the whole cap range, not only at the two caps. The sweep is a separate
$n{=}12$ evaluation, smaller than the main evaluation's $n{=}25$
conditions, so the two give slightly different point estimates at shared caps (e.g.\
$P\!\to\!S@25$ is $0.051$ here vs.\
$0.075$ at $n{=}25$). We always report which $n$ a
number comes from. Per-point bootstrap intervals and per-task scores are in the artifact.

\begin{table}[h]
\centering\small
\caption{Dense cap sweep, marginal means (Opaque Knapsack, $n{=}12$ per cap, the
\emph{same} paired task IDs at every cap). Scores are normalized optimality from a single
temperature-$0.2$ rollout per cap, with no sampling seed. Fig.~\ref{fig:dose}a plots
these with their intervals.}
\label{tab:dense}
\begin{tabular}{lcccccccc}
\toprule
Cap $c$ & 10 & 20 & 25 & 30 & 40 & 50 & 80 & 100 \\
\midrule
$P\!\to\!S$ (mismatch) & 0.00 & 0.00 & 0.05 & 0.05 & 0.23 & 0.23 & 0.64 & 0.55 \\
$P\!\to\!P$ (matched)  & 0.40 & 0.54 & 0.60 & 0.57 & 0.60 & 0.73 & 0.64 & 0.61 \\
\bottomrule
\end{tabular}
\end{table}

\section{Trace-level mechanism analysis}
\label{app:mech}

This appendix works from the recorded trajectories rather than from the scores.
App.~\ref{app:cap} walks through a representative episode in which the cap cuts a planned inspection
sweep and the reset then erases the partial record that sweep was building.
App.~\ref{app:replay} re-executes every recorded episode against the real environment with no model in
the loop, which separates the information a policy gathered from what it did with that information,
and it checks whether the width of the opening block follows the cap the prompt announces or the cap
the runtime enforces.

\subsection{Representative truncation and reset}
\label{app:cap}

One executed action block is one \texttt{run\_code} call. The per-turn cap bounds the number of
tool calls \emph{within} that block. On a cap hit the block stops, the model receives
\texttt{success=False}, the error string \texttt{"Tool call limit exceeded: allowed $N$ per run."}, and
the partial \texttt{stdout} printed \emph{before} the rejected call. Crucially, a cap-rejected
\texttt{inspect} does not decrement the task inspection budget, because the cap check precedes the tool
body. The opening of a real $P\!\to\!S$ episode at $c{=}25$:
\noindent\begin{minipage}{\linewidth}
\begin{lstlisting}
# Turn 0 (assistant code, abbreviated)
all_item_ids = json.loads(list_items())
to_inspect = all_item_ids[:INSPECT_BUDGET]
for item_id in to_inspect:
    data = json.loads(inspect(item_id))   # <-- rejected: 25 total calls (list_items + 24)
# Turn 0 observation:  output "Total items available: 60\n"
#                      error  "Tool call limit exceeded: allowed 25 per run."
# Turn 1 (interpreter has reset; `to_inspect` no longer exists)
batch1 = to_inspect[0:25]
# Turn 1 observation:  error  NameError: name 'to_inspect' is not defined
\end{lstlisting}
\end{minipage}
Building a full list once and reusing it is exactly the persistent-memory strategy training
rewarded. The
binding cap forces that plan across a turn boundary and the reset dismantles it. Complete per-task
rollouts for every reported condition are in the released evaluation traces (App.~\ref{app:repro}).

\subsection{Trace-conditioned replay}
\label{app:replay}

To separate the quality available in the inspected set from the quality the trajectory ultimately
realized, we replay each episode against the real environment with no model in the loop. For each
episode we initialize a fresh replay environment and execute that episode's recorded code blocks in
order, reproducing its runtime semantics and per-turn cap, and we wrap \texttt{inspect} to log every
call. The
re-execution is faithful, since its achieved-over-optimal value reproduces the benchmark's own scores
($0.075/0.662/0.231$
for $P\!\to\!S@25$, $P\!\to\!P@25$,
$P\!\to\!S@40$), so the decomposition describes the real episodes.

Table~\ref{tab:mech} puts the three quantities side by side. \emph{Novel acquisition} first: the
mismatched agent issues $681$ inspect calls per episode but acquires only
$25.7$ distinct items, and its cumulative unique count is nearly flat
after turn~0. Later turns add only
$2.1$ distinct items on average,
while re-listing and re-inspecting the same cached prefix after the reset wipes its progress index. \emph{Coverage vs decision} next: running
the exact solver over $\mathcal{Q}$, the set the policy actually inspected, splits
achieved-over-optimal into coverage $\!=\!\mathrm{OPT}(\mathcal{Q})/\mathrm{OPT}$ and decision
$\!=\!\text{achieved}/\mathrm{OPT}(\mathcal{Q})$. Because $\mathrm{OPT}(\mathcal{Q})$ uses ground-truth
admissibility, coverage is an oracle upper bound over the inspected set and does not imply that the
agent had identified every allowed class represented in $\mathcal{Q}$. The
mismatched inspected set \emph{already contains a $0.71$-of-optimal
solution}, yet the agent captures only $0.09$ of it. The principal binding
constraint is admissibility inference, selection and execution rather than information coverage: the cap is consumed by re-inspection,
leaving no budget to \texttt{take\_item}. The near-threshold recovery at
$c{=}40$ comes almost entirely from improved decision at roughly constant
coverage, not from inspecting more. \emph{Continuation} last: after a truncation the matched agent
continues and the mismatched one starts over.

\begin{table}[h]
\centering\small
\caption{Trace mechanism decomposition (trace-conditioned replay, per-episode means).
\emph{Unique} is the number of distinct items inspected and \emph{replay frac.} the share of inspect
calls that are cached re-inspections. Coverage and decision are defined in the text. \emph{Post-cap restart} is the
fraction of turns following a cap hit that reissue \texttt{list\_items} before acquiring anything new
(App.~\ref{app:diag}), counted over those turns rather than over cap hits, so a truncation the episode
never answers is excluded. Fig.~\ref{fig:gates}b plots the restart column for both runtimes, and it is
the difference between them that carries the continuation claim. The $c{=}40$
row is a replay condition only and has no restart measurement. The cross-task restart values are in
Table~\ref{tab:crosstask}.}
\label{tab:mech}
\setlength{\tabcolsep}{5pt}
\setlength{\tabcolsep}{3.5pt}
\begin{tabular}{lcccccc}
\toprule
Setting & Inspect calls & Unique items & Replay frac. & Coverage & Decision & Post-cap restart \\
\midrule
$P\!\to\!P$, $c{=}25$ (matched)  & 89 & 36.8 & 0.59 & 0.83 & 0.80 & 0.00 \\
$P\!\to\!S$, $c{=}25$ (mismatch) & 681 & 25.7 & 0.96 & 0.71 & 0.09 & \textbf{0.84} \\
$P\!\to\!S$, $c{=}40$      & 660 & 38.2 & 0.94 & 0.81 & 0.33 & --- \\
\bottomrule
\end{tabular}
\end{table}

Context length is a downstream constraint, not the initiating cause. At $T_{\max}{=}40$
a hard context overflow occurs in $3/25$
$P\!\to\!S$ episodes although quality collapses across the entire condition. Overflow becomes common only in
longer or already transcript-heavy runs (App.~\ref{app:tmax}).

\paragraph{Announced versus enforced cap on the opening block.} The cap treatment moves two coupled
channels: the announced \texttt{STRICT LIMIT} line and the enforced runtime limit
(Sec.~\ref{sec:design}). On the \emph{first} action block they separate, because no cap-produced
observation can have reached the model yet. Re-executing that block with the cap removed recovers the
plan the announced number induced, and the two prompts induce the same opening plan: Opaque Knapsack
$P\!\to\!S$ intends $40.2$ calls when
told $25$ and
$40.4$ when told
$80$ (paired difference
$-0.1$,
$[-0.4,+0.2]$),
and Rule Diagnosis intends $4.1$ versus
$4.2$. The Opaque Knapsack agent opens with
the same sweep whichever number it was given, so enforcement is necessary to explain the first
truncation. This concerns the \emph{opening} plan only: once a cap-produced observation exists, the two
channels can no longer be separated.

\section{Cross-task controls and continuation interventions}
\label{app:controls}

This appendix tests the two-condition account of Sec.~\ref{sec:mech} from two directions.
App.~\ref{app:crossfam} varies the task. It specifies the two additional families, Navigation and Rule
Diagnosis, and the bandwidth intervention that narrows Navigation's \texttt{neighbors} interface. The
four settings of Table~\ref{tab:crosstask} include one with frequent truncation but no continuation
advantage, which is what shows that truncation exposure on its own does not produce mismatch-specific
failure. App.~\ref{app:ckpt} varies the runtime. It
specifies the cap-boundary carryover condition, which hands a truncated turn's interpreter variables to
the next turn, and the variant that also announces which variables survived, and it reports what each
repairs and what it leaves unchanged.

\subsection{Navigation and Rule Diagnosis}
\label{app:crossfam}

Sec.~\ref{sec:mech} compares Opaque Knapsack with two further task families, each evaluated
at a single training seed on $16$ instances with its own persistent/stateless adapter
pair (App.~\ref{app:train}). No adapter is run across families or trained on a mixture. The two were
chosen for their distinct \emph{reconstruction channels}: what a reset erases, and how expensively a
tool rebuilds it.

\paragraph{Navigation.} The agent must reach a known \texttt{goal} from a \texttt{start} node in a
directed graph of $n\in[38,73]$ nodes. The edges and node properties
($1$--$2$ traps, $3$--$5$
decoy keys, and $1$ locked gate into the goal whose matching key is reachable
trap-free) are hidden and discovered through tools. Scoring is a phased shortest-path-progress fraction: reaching
the goal scores $1.0$; holding the correct key scores $0.5+0.5\,(1-d/d_{\mathrm{ref}})$ for lock-aware
BFS distance $d$; otherwise $0.5\,(1-d/d_{\mathrm{ref}})$ for progress toward the key.
\texttt{neighbors(nodes)} returns adjacency and is free and \emph{batched} (up to
$50$ nodes per call). \texttt{probe(node)} reveals trap/key/lock properties and
is budgeted and uncached. \texttt{move} and the free validator \texttt{try\_path} complete the API. The
artifact a reset erases is the adjacency map, and batching makes it cheap to rebuild. What matters for
the cap is not that arithmetic but the \emph{width} of the blocks the policy actually emits, meaning
how many tool calls a single block contains, and those blocks are narrow: mean
$5.2$ calls, maximum
$33$.

\paragraph{Rule Diagnosis.} The agent must recover a hidden integer function
$f:[0,999]\to[0,m)$, a contiguous stepwise composition of
interval sub-rules (affine, popcount-affine, or quadratic-mod), with
$m\in\{7,8,9,11\}$, and submits a full-domain hypothesis. We evaluate the
\emph{offline} configuration. \texttt{test\_input(x)} returns $f(x)$ and is budgeted \emph{and
uncached}: each call spends one of $[28,33]$
probes, and re-querying an $x$ re-charges it. \texttt{check(hyp)} is a full-domain equivalence oracle
returning a counterexample, free of the probe budget though it costs a call. A submitted hypothesis
scores $0.60$ functional agreement plus $0.25$ boundary $F_1$ plus
$0.15$ sub-family accuracy, less $0.05$ for each submitted
interval beyond the true interval count, clipped to $[0,1]$. An episode that never submits scores
zero. Because probing is uncached, replaying
a wiped probe map in one block would cost the whole budget
($29.9$ mean, far above the binding cap). The policy never does that: its
widest recorded block is $19$ calls, mean
$3.2$, and not one episode in either condition at either cap
emits a cap-exceeded observation. The nominal budget-to-cap ratio that made this family look
intermediate is simply not the quantity a per-turn cap acts on.

\paragraph{The bandwidth intervention.} We rebuilt the same $16$ Navigation instances
with \texttt{neighbors} limited to $2$ nodes per call. Only that limit changes:
the task JSONs are copies of the evaluated set with a single public field edited, so seeds, graphs,
locks, keys, traps, budgets, scoring and task IDs are identical, and the same adapter and decoding
settings are used. The limit is enforced by the environment and stated in both surfaces the agent reads,
so the cost is planned for rather than discovered through errors. Before running the intervention we
selected $b{=}2$ by re-costing the recorded slack-cap blocks, as the only
candidate predicted to create substantial exposure at $c{=}25$ while remaining
slack at $c{=}80$. The selection procedure and the per-candidate calculations are
included with the artifact. It is a design calculation, not evidence: it holds the recorded code fixed,
whereas an agent facing the narrower interface writes different code.

\paragraph{What the four settings say.} Table~\ref{tab:crosstask} is the comparison. Limiting
\texttt{neighbors} to $2$ nodes per call makes the agent spend many more calls to
map the same graph. Its blocks widen from
$5.2$ to
$21.4$ calls on average, and the cap-hit rate rises from
$4\%$ to
$87\%$. Truncation is now routine, which satisfies
the first of the two conditions in Sec.~\ref{sec:mech}: the cap must cut an action the agent intended
to finish.

Quality then falls in \emph{both} runtimes as the cap tightens from
$c{=}80$ to $c{=}25$, from
$0.44$ to $0.04$ under
mismatched deployment and from $0.48$ to
$0.17$ under matched deployment. Cap amplification of the runtime
gap, $G_P(25)-G_P(80)$, is
$+0.08$ under the narrowed interface against
$-0.17$ under the original one. What the intervention estimates is the
difference between those two arms, $+0.25$, the direction Opaque Knapsack
predicts, but its interval spans zero
($[-0.18,+0.70]$,
$n{=}16$), so we do not detect mismatch-specific amplification here.

The restart rates are the behavior to read alongside it. After a truncation both agents frequently
rebuild the map from the beginning instead of carrying on from what survived. That is consistent with
little continuation advantage under the narrowed interface, while the outcome interval stays too wide
to rule out additional mismatch-specific amplification. Frequent truncation on its own therefore does
not single out the mismatched runtime here. Run-to-run variation is a second reason to read the
$+0.25$ difference cautiously. Re-running the archived control condition with
nothing changed moved its score by $+0.10$
($[-0.02,+0.26]$), an interval that reaches
the size of the difference we would be claiming, so we do not read a secondary-task difference of this
size as established.

\begin{table}[h]
\centering\small
\caption{Cross-task exposure and continuation. $E_{25}$ is the fraction of
slack-cap action blocks whose replayed call count exceeds the binding cap. Restart is the post-cap
re-derivation fraction of Table~\ref{tab:mech}. Plain Navigation has too few post-cap turns to estimate
either restart fraction ($n{=}6$ and
$11$), and Rule Diagnosis has none at all.}
\label{tab:crosstask}
\setlength{\tabcolsep}{3.5pt}
\begin{tabular}{p{0.15\columnwidth} p{0.20\columnwidth} c c c p{0.21\columnwidth}}
\toprule
Setting & Tool organization & $E_{25}$ & Restart $P\!\to\!P$ & Restart $P\!\to\!S$ & Interpretation \\
\midrule
Opaque Knapsack & Long batched inspection & 0.53 & 0.00 & 0.84 & Mismatch-specific failure \\
Navigation & Broad batched topology & 0.04 & Too few events & Too few events & Cap rarely exposes mismatch \\
Navigation $b{=}2$ & Narrow topology batches & 0.42 & 0.98 & 0.94 & Both runtimes deteriorate \\
Rule Diagnosis & Short adaptive probe cycles & 0.00 & --- & --- & No exposure \\
\bottomrule
\end{tabular}
\end{table}

\paragraph{Control-task outcomes.} Table~\ref{tab:crosstaskq} gives the quality and the cost behind the
reading above, for the full train$\times$runtime factorial at both caps, and
Table~\ref{tab:crosstaskamp} gives the paired cap amplification of each runtime gap. Both controls are
one seed, one rollout and $n{=}16$ paired instances, so every interval here is wide. Read them as
showing what was measured rather than as bounding an effect.

Three things are visible. First, neither control gives statistically resolved evidence of positive cap
amplification in the persistent agent's runtime gap: the estimate is
$-0.17$ on Navigation and $+0.11$ on Rule Diagnosis, with both
intervals spanning zero. Second, the quality means are away from the floor and the ceiling in every
cell, although that does not rule out saturation on individual instances. Third, the intervals are wide enough to
be compatible with a moderate response in either direction, which is why we report that we did not
detect one rather than that none exists. The stateless-trained rows are the least constrained of all:
their amplifications, $+0.32$ and $-0.14$, have intervals wider
than either point estimate.

The cost columns qualify the claim that mismatch is paid for in tokens. On Navigation the mismatched
condition spends more than matched at both caps, roughly $2.4\times$ at each. On Rule Diagnosis it
spends more at the slack cap and \emph{less} at the binding cap, because its mismatched episodes are
shorter there, so repeated work does not always show up as a larger token bill.

\begin{table}[h]
\centering\small
\caption{Control-task outcomes: quality and cost for the full train$\times$runtime factorial at both
caps ($n{=}16$ paired instances per cell, one seed, one rollout). Quality is each family's own score,
phased shortest-path progress for Navigation and, for Rule Diagnosis, the composite rule score defined
in App.~\ref{app:crossfam}. Tokens
are per-episode means of recorded usage for accepted completions.}
\label{tab:crosstaskq}
\begin{tabular}{llcccc}
\toprule
Task & Condition & $Q(c{=}25)$ & $Q(c{=}80)$ & Tokens @$25$ & Tokens @$80$ \\
\midrule
\multirow{4}{*}{Navigation}
 & $P\!\to\!P$ (matched)  & 0.55 & 0.68 & 142k & 155k \\
 & $P\!\to\!S$ (mismatch) & 0.41 & 0.38 & 352k & 366k \\
 & $S\!\to\!P$            & 0.32 & 0.44 & 292k & 228k \\
 & $S\!\to\!S$ (matched)  & 0.53 & 0.33 & 200k & 256k \\
\midrule
\multirow{4}{*}{Rule Diagnosis}
 & $P\!\to\!P$ (matched)  & 0.70 & 0.69 & 102k & 61k \\
 & $P\!\to\!S$ (mismatch) & 0.33 & 0.41 & 77k  & 108k \\
 & $S\!\to\!P$            & 0.34 & 0.22 & 130k & 136k \\
 & $S\!\to\!S$ (matched)  & 0.18 & 0.21 & 133k & 178k \\
\bottomrule
\end{tabular}
\end{table}

\begin{table}[h]
\centering\small
\caption{Cap amplification of each runtime gap on the control tasks, over the same
$16$ paired instances. $G_P(c){=}Q_{PP}(c)-Q_{PS}(c)$ is the persistent-trained agent's runtime gap and
$G_S(c){=}Q_{SS}(c)-Q_{SP}(c)$ is the stateless-trained agent's. Amplification is the paired difference
$G(25)-G(80)$, with a $95\%$ bootstrap CI over task instances. Every interval spans zero.}
\label{tab:crosstaskamp}
\setlength{\tabcolsep}{4pt}
\begin{tabular}{lccc@{\quad}ccc}
\toprule
Task & $G_P(25)$ & $G_P(80)$ & $A_P$ & $G_S(25)$ & $G_S(80)$ & $A_S$ \\
\midrule
Navigation     & $+0.13$ & $+0.30$ & \begin{tabular}[t]{@{}c@{}}$-0.17$\\[-1pt]{\scriptsize $[-0.50,+0.15]$}\end{tabular} & $+0.21$ & $-0.11$ & \begin{tabular}[t]{@{}c@{}}$+0.32$\\[-1pt]{\scriptsize $[-0.12,+0.81]$}\end{tabular} \\
\addlinespace[2pt]
Rule Diagnosis & $+0.38$ & $+0.27$ & \begin{tabular}[t]{@{}c@{}}$+0.11$\\[-1pt]{\scriptsize $[-0.17,+0.39]$}\end{tabular} & $-0.16$ & $-0.01$ & \begin{tabular}[t]{@{}c@{}}$-0.14$\\[-1pt]{\scriptsize $[-0.38,+0.10]$}\end{tabular} \\
\bottomrule
\end{tabular}
\end{table}

\subsection{Cap-boundary carryover: the continuation intervention}
\label{app:ckpt}

\paragraph{The condition.} A third runtime, between the two the paper otherwise uses. An ordinary turn is
followed by the usual reset. A turn terminated by \texttt{Tool call limit exceeded} has its globals
handed to the next turn, and to that turn only. Consecutive truncations chain. What the condition does
\emph{not} restore bounds what a null result supports: the cap aborts the block mid-execution, so the
instruction pointer is gone and the interrupted loop does not continue, and there is no explicit
next-index checkpoint and no directive telling the policy where to pick up. This is the smallest patch
addressing the boundary the trace analysis implicates, not the largest available. Automatic
continuation, injected checkpoint records and explicit resumption metadata are runtime repairs this
paper does not test.

Held fixed against $P\!\to\!S$: adapter, instances, per-turn cap, turn budget, timeouts, temperature,
the reset system prompt and few-shot examples, the cap error text, and the \texttt{runtime\_state}
banner, which still reports \texttt{reset} with an empty \texttt{active\_globals}, so the agent is
never told its working record survived. Three conditions ($P\!\to\!S$, $P\!\to\!S$ with carryover, and
$P\!\to\!P$) ran over the same $25$ instances at $c{=}25$,
each task seeded from a common per-task decoding seed, all three re-run together because the rescue
fraction divides one condition difference by another. The design, quantity, bootstrap, secondary panel and
the reading of each outcome range (including the reading of a null) were fixed before any episode ran,
though not through a third-party registration. The manipulation held: the carryover condition hits the cap on
$65\%$ of executed turns and carries a checkpoint
into $22.2$ turns per episode, and the announced
variant named the survivors on $100\%$ of
truncated turns against $0\%$ silent.
Table~\ref{tab:ckptfull} gives the pre-specified panel.

\begin{table}[h]
\centering\small
\caption{Cap-boundary carryover evaluations, with trace-derived behavioral metrics
($n{=}25$ per condition). The
first three conditions are one batch and the last two another, so read quality down each block, not across
them. ``Carried'' turns are those following a truncated one. In conditions without carryover those are the
turns a checkpoint \emph{would} have been carried into.}
\label{tab:ckptfull}
\begin{tabular}{lccccc}
\toprule
Quantity & $P\!\to\!S$ & $+$Carry & $P\!\to\!P$ & Carry (re-run) & $+$Announced\\
\midrule
Quality & 0.038 & 0.132 & 0.575 & 0.081 & 0.080\\
\texttt{NameError} turns/ep & 9.6 & 1.3 & 0.1 & 6.0 & 5.0\\
\dots{} of them on carried turns & 6.2 & 0.0 & 0.1 & 0.0 & 0.0\\
Re-derives after a truncation & 0.76 & 0.76 & 0.00 & 0.86 & 0.31\\
Distinct items & 24.2 & 28.2 & 34.1 & 28.0 & 29.6\\
Decision achieved/OPT($\mathcal{Q}$) & 0.08 & 0.23 & 0.72 & 0.11 & 0.11\\
Ends on $T_{\max}$ & 96\% & 80\% & 12\% & 92\% & 92\%\\
\bottomrule
\end{tabular}
\end{table}

\paragraph{Reading.} Silent carryover eliminates the missing-variable errors it targets but closes only
$17\%$ of the quality gap
($[1,36]\%$) and does not reduce
restarting. Announcing the surviving variables does reduce restarting, with re-derivation falling from
$0.86$ to
$0.31$ of post-cap turns, but it leaves
quality unchanged ($-0.001$
$[-0.129,+0.121]$).
State availability and visibility are therefore insufficient without a policy that resumes productive
computation. Two bounds on that reading. The announced banner names live bindings while the runtime
banner still says \texttt{reset} and the few-shot examples still demonstrate rebuilding, so the policy
receives a mixed signal. A null says naming the survivors is insufficient \emph{under an otherwise
reset-oriented prompt and policy package}, not that a coherently informed or retrained policy could not
use what survived. And the gap-closure fraction is descriptive, not an identified causal mediation
proportion: identifying one would need assumptions about the mediator that a three-condition runtime comparison
neither delivers nor was designed to test. The outcome effects also stay imprecise: each condition is one
rollout per task at $n{=}25$, and re-running the silent condition with seeds held fixed
moved its mean by $-0.051$ between serving batches.

\section{Robustness and replication}
\label{app:robust}

\subsection{Episode horizon and context}
\label{app:tmax}

Table~\ref{tab:tmax} asks whether the binding-cap collapse is merely an artifact of too short an
episode. Quadrupling the horizon leaves the mismatched condition collapsed with no upward trend. The
paired changes are
$-0.018$ $[-0.136,+0.107]$ and
$+0.012$ $[-0.128,+0.156]$,
small and straddling zero, while the matched condition remains stable at every horizon. This is a joint
turn-and-time-budget increase, not a pure horizon control, and the wall-clock ceiling is
\emph{reached} for the mismatched condition: $13/20$ and
$17/20$ of its episodes terminate in timeout against
$2/20$ and $4/20$ matched. The
added turns and the added wall-clock are absorbed by repeated state re-emission until the clock or the
serving window is reached. So the claim is narrow: the default turn cutoff does not explain the collapse,
and
raising the practical turn \emph{and} time allowance does not rescue the observed recovery policy, on
instances the matched agent solves within the same budget. It is \emph{not} a claim that recovery is
impossible under arbitrarily large time and context limits. Unbounded-horizon deployment is out of scope
(Sec.~\ref{sec:limitations}).

\begin{table}[h]
\centering\small
\caption{Episode-horizon sweep at $c{=}25$ (Opaque Knapsack, normalized
optimality), on the same common set of $20$ instances, with the per-turn context
behavior that accompanies it. The $80/160$ configs
raise both the turn limit $T_{\max}$ and the episode wall-clock
($3{,}000\!\to\!4{,}000$\,s). ``Overflow'' counts episodes with at least one
context-window overflow. The runtime hard-errors rather than silently truncating once a
turn's input plus the $8192$-token output reservation would exceed the
$40960$ window. Overflow denominators are each condition's own full episode set, not the
common set the quality rows use.}
\label{tab:tmax}
\begin{tabular}{lccc}
\toprule
& $T_{\max}{=}40$ & $T_{\max}{=}80$ & $T_{\max}{=}160$ \\
\midrule
$P\!\to\!P$ (matched)  & $0.66$  & $0.59$  & $0.66$  \\
$P\!\to\!S$ (mismatch) & $0.076$ & $0.058$ & $0.088$ \\
\midrule
$P\!\to\!S$ overflow & $3/25$ & $13/20$ & $13/20$ \\
$P\!\to\!P$ overflow & $4/25$ & --- & --- \\
\bottomrule
\end{tabular}
\end{table}

\paragraph{What the horizon counts.} $T_{\max}$ bounds the model calls an episode may attempt, not the
action blocks it executes. A call that raises, and a completion the harness rejects because the model
did not stop on its own, each spend one attempt, and neither logs a step nor reaches the token counter.
Table~\ref{tab:horizon} separates the three quantities for the main evaluation. Attempted turns exceed
logged steps wherever the serving window overflows, which is how the mismatched condition averages
$35.2$ logged steps while $23$ of its $25$ episodes end at a horizon of $40$: its remaining attempts are
context-window rejections. Attempted is the sum of the three, so the check that carries information is whether it lands on
the horizon: every episode that ended at the turn limit accounts for all $40$ attempts exactly, in
all $77$ such episodes here and in all $714$ across the archive. Executed action blocks equal logged steps in
every cell of this evaluation, since no step logged an empty block. This accounting is read off the
released harness, which is the evaluated one on this code path: the turn loop and both of its
early-exit branches last changed on 2026-02-14, and the earliest of the archived episodes began on
2026-06-21.

The token column of Table~\ref{tab:main2x2} is the usage recorded for accepted completions. Two
categories sit outside it. Most failed calls are context-window rejections, which the provider refuses
before generating anything, so they consume no output tokens. The remaining $13$ are transport failures
after six retries, $12$ gateway timeouts and one connection error, and a failed request is not by
itself evidence that nothing was generated: their usage never returned to the client and is
unavailable. All $13$ fall in $S\!\to\!P$ and $S\!\to\!S$. In the two $c{=}80$ conditions behind the
$4.4\times$ ratio, every failed call was a context-window rejection. Rejected completions do carry
generated tokens, and they are the ones worth auditing: there are none anywhere in this evaluation.
Across the $1{,}974$ archived episodes behind the paper there are seven in total, all in the
replication arms of App.~\ref{app:arms}, which report quality and not token totals. No returned
completion was excluded from the token totals in the conditions used for a reported token
comparison.

\begin{table}[h]
\centering\small
\caption{Horizon accounting for the main evaluation (Opaque Knapsack, seed-0,
$n{=}25$ episodes per condition, $T_{\max}{=}40$). \emph{Attempted} and \emph{Steps} are per-episode
means of model calls attempted and steps logged. \emph{Failed} and \emph{Rejected} are totals over the
$25$ episodes, the first counting calls that raised and the second counting completions rejected for
not stopping on their own. \emph{Context} is the share of the failed calls that were serving-window
rejections. Termination counts are in Table~\ref{tab:main2x2}, and every episode that ended at the
horizon satisfies attempted $=$ steps $+$ failed $+$ rejected.}
\label{tab:horizon}
\begin{tabular}{llccccc}
\toprule
Cap & Condition & Attempted & Steps & Failed & Context & Rejected \\
\midrule
\multirow{4}{*}{$c{=}25$}
 & $P\!\to\!P$ (matched)  & 13.2 & 9.8  & 85  & 85/85   & 0 \\
 & $P\!\to\!S$ (mismatch) & 37.3 & 35.2 & 52  & 52/52   & 0 \\
 & $S\!\to\!P$            & 27.8 & 18.6 & 229 & 227/229 & 0 \\
 & $S\!\to\!S$ (matched)  & 21.6 & 11.2 & 260 & 260/260 & 0 \\
\midrule
\multirow{4}{*}{$c{=}80$}
 & $P\!\to\!P$            & 4.6  & 4.6  & 0   & 0/0     & 0 \\
 & $P\!\to\!S$            & 14.5 & 13.0 & 37  & 37/37   & 0 \\
 & $S\!\to\!P$            & 14.0 & 10.0 & 101 & 93/101  & 0 \\
 & $S\!\to\!S$            & 16.5 & 9.8  & 169 & 166/169 & 0 \\
\bottomrule
\end{tabular}
\end{table}

\subsection{Replications: a second rollout and two base models}
\label{app:arms}

Every replication is the same Opaque Knapsack $2\times2\times\{25,80\}$
experiment design (two training runtimes, two deployment runtimes and the two per-turn caps) over the
same $25$ instances, so each quantity is paired on the instance
and bootstrapped over tasks ($5000$ resamples,
$95\%$ percentile intervals).
Table~\ref{tab:armsfull} reports them.

\paragraph{Scope of the replication.} That factorial is all these replications repeat. Rule Diagnosis, both
Navigation variants, the trace analysis of Sec.~\ref{sec:mech}, and the carryover conditions of
App.~\ref{app:ckpt} were run on the main Qwen3-8B setting only. What they support is
therefore that the primary Opaque Knapsack runtime gap and its cap amplification are not artifacts of one base
checkpoint or one sampling run. They do not show that the mechanism analysis or the boundary
interventions have been reproduced elsewhere.

\paragraph{What each replication holds fixed, and what it cannot.} The second rollout re-runs the seed-0
adapters at the same temperature with fresh sampling and no decoding seed: adapters, instances, prompts,
caps and turn budget are identical, so the only source of difference is decoding and any serving
nondeterminism. Across the second rollout the mean absolute condition shift is
$0.057$ (maximum $0.159$) and no
condition-level difference excludes zero. The SD of the \emph{per-task} differences is several times larger
than any condition mean moves, which is why every quantity in this paper is paired on the instance rather
than compared as independent means. The two base-model replications retrain the same LoRA recipe on a different
base checkpoint and are evaluated through identical benchmark configs, but two differences were forced
by those checkpoints and are not controlled. Mistral-7B-v0.3 serves at its architectural maximum of $32$k context
against $40$k, which matters
most for the mismatched condition, whose prompt grows fastest. And neither base checkpoint ships a chat
template, so each replication uses a hand-written one, applied identically at training and serving time.
Mistral's has no system role, since it folds system messages into the following user turn, so the
runtime banner, few-shot examples and tool prompt reach that model inside a user turn rather than as
system messages.

\paragraph{The model-selection difference is secondary.} Under the same stateless deployment runtime, the
cap also moves the point estimate of which \emph{training} runtime scores better. On the full seed-0
cohort the stateless-trained agent scores $0.19$
$[0.05,0.31]$ higher at the
binding cap, and that advantage shrinks by $0.30$
$[0.11,0.49]$ at the slack cap, where the
point estimate reverses. That apparent sign reversal is sensitive to stateless-condition timeouts.
Dropping the $5$ timed-out slack-cap episodes removes it, and on the
timeout-free paired cohort the stateless-trained agent stays ahead at both caps. It also does not
reproduce in the second rollout or in either base-model replication. We therefore interpret only the
moderation of the gap, never a ranking flip.

\begin{table}[h]
\centering\small
\caption{Replications. $G_P$ is the persistent model's runtime gap and
$G_P(25)-G_P(80)$ its cap amplification. $T$ is the change in
the train$\times$runtime interaction across the caps, and $M=Q_{PS}-Q_{SS}$ is the secondary model-selection
difference under stateless deployment. The runtime gap and its amplification are present within every
replication.
$M(80)$ is the quantity whose \emph{sign} the main text declines to claim: it is
never positive with an interval that excludes zero, and is reliably negative in the two base-model settings.}
\label{tab:armsfull}
\setlength{\tabcolsep}{4pt}
\begin{tabular}{lcccc}
\toprule
 & Qwen3-8B & +2nd rollout & Mistral-7B-v0.3 & Llama-3.1-8B\\
\midrule
$G_P(25)$ & +0.59 [+0.46,+0.71] & +0.63 [+0.50,+0.75] & +0.64 [+0.50,+0.78] & +0.58 [+0.43,+0.72]\\
$G_P(80)$ & +0.15 [+0.00,+0.32] & +0.08 [-0.06,+0.21] & +0.14 [-0.01,+0.28] & +0.18 [+0.01,+0.35]\\
Cap amplification & +0.44 [+0.22,+0.65] & +0.55 [+0.38,+0.71] & +0.51 [+0.32,+0.71] & +0.40 [+0.16,+0.63]\\
$M(80)$ & +0.11 [-0.08,+0.29] & -0.02 [-0.15,+0.09] & -0.18 [-0.32,-0.04] & -0.23 [-0.42,-0.04]\\
$T$ & +0.31 [+0.04,+0.57] & +0.40 [+0.24,+0.55] & +0.41 [+0.09,+0.74] & +0.34 [+0.03,+0.65]\\
\bottomrule
\end{tabular}
\end{table}

\section{Artifacts and reproducibility}
\label{app:repro}

Table~\ref{tab:artifacts} lists the released artifacts. The code is at
\url{https://github.com/TieuDaoChanNhan/runtime-contract}. It documents the directory layout, the per-condition configurations and the
regeneration commands. The evaluation data is at
\url{https://huggingface.co/datasets/runtime-contracts/evaluation-traces} and the teacher training
traces are at
\url{https://huggingface.co/datasets/runtime-contracts/teacher-traces}. All 14 trained adapters are
on Hugging Face pages under the same organization.

\begin{table}[h]
\centering\small
\caption{Released artifacts.}
\label{tab:artifacts}
\begin{tabular}{p{0.19\columnwidth} c p{0.60\columnwidth}}
\toprule
Artifact & Released & Contents \\
\midrule
Code & Yes & Task generators, both runtimes, agent harness, training and serving scripts, every condition's benchmark config, analysis scripts, and the verbatim prompts \\
Evaluation traces & Yes & Per-task rollouts of every reported condition: main evaluation, dense sweep, horizon control, cross-task controls, both interventions, replications \\
Opaque Knapsack adapters & Yes & 6 Qwen3-8B (persistent/stateless $\times$ 3 seeds) $+$ 4 replication adapters, one card each \\
Control-task adapters & Yes & The 4 Navigation and Rule Diagnosis adapters, same recipe and card format \\
Teacher traces & Yes & The filtered, instance-paired SFT data of App.~\ref{app:train}, as fed to the fine-tuning script. The raw cached teacher responses it was built from are not released \\
\bottomrule
\end{tabular}
\end{table}

Regenerating the reported numbers needs no GPU and no new rollouts: with the evaluation data in place,
one analysis script recomputes every cited number into a single JSON file. What a re-run does \emph{not} reproduce is the digits:
evaluation decodes at temperature $0.2$ with no decoding seed and serving is
not bit-deterministic, so the released traces, not a re-run, are what fixes the reported values,
and App.~\ref{app:arms} measures how far a re-run moves them. The same split applies on the training
side: the released teacher traces are the processed data as fed to the fine-tuning script, so they fix the
training data exactly, but the raw cached teacher responses they were built from are not released and teacher
decoding was unseeded, so the original stochastic teacher generation is not independently reproducible.

Everything the paper reports is covered in full, across two artifacts rather than one. The code
repository carries the configurations and the analysis scripts. The evaluation dataset carries the
per-task results and traces, for every condition behind a reported number. After downloading the
dataset into the directory layout the repository documents, a reader recomputes each reported value
from its episodes rather than reading it off a summary, with no new model rollouts, and the bootstrap
intervals of the dense cap sweep, both interventions and the replications reproduce along with their
point estimates. The release is pinned by revision: the code is commit \texttt{230d4840}, tagged
\texttt{arxiv-v1}, the evaluation traces are revision \texttt{378805a6} and the teacher traces
revision \texttt{597f91c6}. All three are readable without an account. On licenses: Qwen3-8B and
Mistral-7B-v0.3 are Apache-2.0 licensed, Llama-3.1-8B is released under the
Llama 3.1 Community License (a custom license rather than an OSI-approved open-source one,
used here under its research and acceptable-use terms), and this paper's own code is
Apache-2.0 licensed.

\section{Broader impact}
\label{app:impact}

Evaluating at a single favorable resource setting can hide failures that appear under tighter
constraints, and can materially change the apparent performance gap between trained agents
(Sec.~\ref{sec:main}). In our primary rollout the point-estimate ranking also changes, although that
sign does not replicate (App.~\ref{app:arms}). The robust finding is that the magnitude of the gap
depends strongly on the cap. Reporting the runtime contract and varying the cap, as
Sec.~\ref{sec:protocol} recommends, mitigates that reporting risk. We report both a slack and a
binding configuration throughout.

The societal impacts run in both directions. More reliable runtime-sensitive evaluation may reduce
hidden deployment failures and unnecessary inference cost. Conversely, evaluating multiple runtime and
cap configurations increases experimental compute, and selective reporting of favorable configurations
could still be used to overstate agent reliability.

Because our evidence comes from procedurally generated tasks and 7--8B fine-tuned agents, it should be
read as guidance for evaluation methodology rather than as a direct characterization of deployed
systems.

\section{Notation and terminology}
\label{app:notation}

Tables~\ref{tab:notation} and~\ref{tab:notation2} collect the symbols and terms the paper uses, with
the section, table or appendix where each is introduced.

\begin{table}[t]
\caption{Notation and terminology: runtimes, conditions and evaluation settings.}
\label{tab:notation}
\centering\small
\begin{tabular}{@{}L{0.21\linewidth}p{0.73\linewidth}@{}}
\toprule
\multicolumn{2}{@{}l}{\textit{Runtime contracts and conditions}} \\
\midrule
$P$, $S$ & The \emph{persistent} runtime, under which the variables an agent creates stay available on later turns, and the \emph{stateless} runtime, under which each turn begins with a fresh set of them. Task and environment state, including the tool objects, survives under both (Sec.~\ref{sec:design}) \\
Runtime contract & The interpreter semantics alone: whether agent-created bindings survive the turn boundary (Sec.~\ref{sec:design}) \\
Runtime condition & The contract together with the prompt package that announces it: a mode instruction, a runtime-state banner and runtime-specific few-shot examples. Cross-runtime comparisons switch the whole package, not the semantics in isolation (Sec.~\ref{sec:design}, App.~\ref{app:runtime}) \\
$X\!\to\!Y$ & An $X$-trained agent deployed on a $Y$ runtime. $P\!\to\!P$ and $S\!\to\!S$ are the matched references, $P\!\to\!S$ the main mismatch and $S\!\to\!P$ the reverse transfer (Table~\ref{tab:design}) \\
Condition & One of the four $X\!\to\!Y$ combinations, evaluated at one cap (Table~\ref{tab:design}) \\
Runtime transfer & Whether an agent trained under one runtime still solves the task when deployed under the other (Sec.~\ref{sec:intro}) \\
Working record & The task information an agent accumulates from tool feedback and reuses in later actions: the inspected-item table, the map of locks, traps and keys, or the probes and hypothesis (Sec.~\ref{sec:intro}, Table~\ref{tab:tasks}) \\
Replication & A repeat of the Opaque Knapsack factorial with a different rollout or base checkpoint: two Qwen3-8B rollouts, Mistral-7B-v0.3 and Llama-3.1-8B (Fig.~\ref{fig:arms}, App.~\ref{app:arms}) \\
Silent carryover & An intervention runtime that resets every turn \emph{except} one the cap truncated, whose variables survive into the next turn only. Its \emph{announced} variant also names the survivors in the observation (Sec.~\ref{sec:mech}, App.~\ref{app:ckpt}) \\
\addlinespace
\multicolumn{2}{@{}l}{\textit{Evaluation settings}} \\
\midrule
Action block, turn & The Python block a cap truncates, and the unit the episode horizon counts. A turn is one attempted model call and executes at most one action block, so a failed or rejected call spends a turn without executing one (Sec.~\ref{sec:design}, App.~Table~\ref{tab:horizon}) \\
$c$ & Per-turn cap: the tool calls allowed inside one action block. On reaching it the runtime stops execution before the next call, and the rejected call does not consume the task budget (Sec.~\ref{sec:design}, Table~\ref{tab:limits}) \\
Slack, binding & Labels for the intended cap regimes, $c{=}80$ and $c{=}25$, preselected from task mechanics. Whether a cap is reached in practice is measured by the cap-hit rate and the exposure (Sec.~\ref{sec:design}) \\
$T_{\max}$ & Episode turn limit, $40$ ($80$ and $160$ in the horizon sweep). Distinct from cap moderation $T$ below (Table~\ref{tab:limits}, App.~\ref{app:tmax}) \\
$B$ & Opaque Knapsack's inspection budget, counting \emph{distinct} item IDs: re-inspecting a seen ID returns the cached record and consumes no further inspection budget, though it still spends a call from the per-turn cap (App.~\ref{app:task}). The Navigation and Rule Diagnosis probe budgets are \emph{uncached}, so a repeated probe is charged again (App.~\ref{app:crossfam}) \\
$b$ & Navigation batch limit: nodes returned per \texttt{neighbors} call, cut to $b{=}2$ in the narrowed control (Sec.~\ref{sec:mech}, App.~\ref{app:crossfam}) \\
$n$ & Paired task instances per condition: $25$ in the main evaluation, $12$ in the dense cap sweep, $16$ in the cross-task controls, $8$ in the three-seed sign replication, $20$ in the horizon sweep and $25$ in the carryover interventions \\
\bottomrule
\end{tabular}
\end{table}

\begin{table}[t]
\caption{Notation and terminology: the outcome, the reported quantities and the trace-derived ones.}
\label{tab:notation2}
\centering\small
\begin{tabular}{@{}L{0.21\linewidth}p{0.73\linewidth}@{}}
\toprule
\multicolumn{2}{@{}l}{\textit{Outcome and reported quantities}} \\
\midrule
Normalized optimality & The outcome: achieved value over the exact optimum on allowed-class items, clamped to $0$ on any capacity overflow (Sec.~\ref{sec:design}, App.~\ref{app:task}) \\
$Q_{XY}(c)$ & Mean task quality for an adapter trained under runtime $X$ and evaluated under runtime $Y$ at cap $c$: normalized optimality for Opaque Knapsack, and each control family's own score (Sec.~\ref{sec:design}, App.~\ref{app:crossfam}) \\
$G_P(c)$ & Runtime gap: $Q_{PP}(c)-Q_{PS}(c)$ (Sec.~\ref{sec:design}) \\
$D(c)$ & Train$\times$runtime interaction: $\bigl(Q_{PP}(c)-Q_{PS}(c)\bigr)-\bigl(Q_{SP}(c)-Q_{SS}(c)\bigr)$ (Sec.~\ref{sec:design}) \\
$T$ & Cap moderation: $D(25)-D(80)$ (Sec.~\ref{sec:design}) \\
Cap amplification & The same tightening applied to the runtime gap itself: $G_P(25)-G_P(80)$ (Sec.~\ref{sec:design}, Fig.~\ref{fig:arms}) \\
$M(c)$ & Secondary model-selection difference under stateless deployment: $Q_{PS}(c)-Q_{SS}(c)$ (App.~\ref{app:arms}) \\
$\beta$ & Slope of quality on $\log c$ in the task-panel model $Q_{ic}=\alpha_i+\beta\log c+\epsilon_{ic}$, with a fixed effect per task ID (Sec.~\ref{sec:main}, App.~\ref{app:dense}) \\
\addlinespace
\multicolumn{2}{@{}l}{\textit{Trace-derived quantities} (defined in Table~\ref{tab:diag})} \\
\midrule
Cap-hit rate & Mean per-episode fraction of logged steps whose interpreter error reports the per-turn tool-call limit (Sec.~\ref{sec:design}, Table~\ref{tab:diag}) \\
Truncation exposure $E_c$ & Fraction of a condition's slack-cap action blocks whose replayed call count exceeds $c$ (Sec.~\ref{sec:mech}, App.~\ref{app:diag}) \\
Post-cap restart & The next executed turn after a cap hit reissues the task's listing or index call before acquiring anything new (Sec.~\ref{sec:mech}, Table~\ref{tab:diag}). Reserved for this conditional rate, and not used for the unconditional listing-call count of Table~\ref{tab:main2x2} \\
Continuation advantage & The persistent runtime lets the matched agent continue from partial state where the stateless runtime makes the mismatched agent restart (Sec.~\ref{sec:mech}) \\
\texttt{NameError} count & Logged steps raising a missing-variable error, per episode: the signature of a reset destroying a binding the policy assumed persisted (Sec.~\ref{sec:main}, Table~\ref{tab:diag}) \\
Coverage, decision & $\mathrm{OPT}(\mathcal{Q})/\mathrm{OPT}$ and achieved$/\mathrm{OPT}(\mathcal{Q})$, where $\mathcal{Q}$ is the set of items the agent actually inspected (App.~\ref{app:replay}) \\
\bottomrule
\end{tabular}
\end{table}

\clearpage


\end{document}